\pdfoutput=1
\documentclass[times]{elsarticle}
\makeatletter
\def\ps@pprintTitle{%
	\let\@oddhead\@empty
	\let\@evenhead\@empty
	\def\@oddfoot{\centerline{\thepage}}%
	\let\@evenfoot\@oddfoot}
\makeatother
\usepackage{epsfig}
\usepackage{graphicx}
\usepackage{amsmath}
\usepackage{amssymb}
\usepackage{algorithm} % algorithm
\usepackage{algpseudocode}
\usepackage{booktabs}

\usepackage{caption}
\usepackage{subcaption}
\usepackage{color}

\usepackage{picins}
\usepackage{wrapfig}
\usepackage{times}
\usepackage{amssymb}
\usepackage{helvet}
\usepackage{mathrsfs}
\usepackage{url}
\usepackage{courier}
\usepackage{booktabs}
\usepackage{threeparttable}

\usepackage{url}	
\usepackage{overpic}
\usepackage{color}
\usepackage[english]{babel}
\usepackage[T1]{fontenc}
\usepackage[utf8]{inputenc}
\usepackage{multirow}
\usepackage{diagbox}
\usepackage{algpseudocode}

\biboptions{sort&compress}

\journal{Neurocomputing}

\begin{document}

\begin{frontmatter}



\title{Revisiting Metric Learning for Few-Shot Image Classification}

\author{Xiaomeng Li$^{1}$,Lequan Yu$^{1}$,Chi-Wing Fu$^{1}$, Meng Fang$^{3}$, Pheng-Ann Heng$^{1,2}$} % Replace with your names

\address{$^1$Department of Computer Science and Engineering, The Chinese University of Hong Kong \\
$^2$ Shenzhen Key Laboratory of Virtual Reality and Human Interaction Technology, Shenzhen Institutes of Advanced Technology, Chinese Academy of Sciences, China \\
$^3$ Tencent AI Lab}

\begin{abstract}
The goal of few-shot learning is to recognize new visual concepts with just a few amount of labeled samples in each class.
Recent effective metric-based few-shot approaches employ neural networks to learn a feature similarity comparison between query and support examples.
However, the importance of feature embedding, $i.e.$, exploring the relationship among training samples, is neglected.
In this work, we present a simple yet powerful baseline for few-shot classification by emphasizing the importance of feature embedding.
Specifically, we revisit the classical triplet network from deep metric learning, and extend it into a deep $K$-tuplet network for few-shot learning, utilizing the relationship among the input samples to learn a general representation learning via episode-training.
Once trained, our network is able to extract discriminative features for unseen novel categories and can be seamlessly incorporated with a non-linear distance metric function to facilitate the few-shot classification.
Our result on the miniImageNet benchmark outperforms other metric-based few-shot classification methods.
More importantly, when evaluated on completely different datasets (Caltech-101, CUB-200, Stanford Dogs and Cars) using the model trained with miniImageNet, our method significantly outperforms prior methods, demonstrating its superior capability to generalize to unseen classes.

\if 1 
%The well-trained feature embedding network can be seamlessly incorporated with a non-linear distance metric function to extract discriminative features for unseen category recognition.
%Once trained, it is able to extract discriminative features for unseen novel categories to facilitate the few-shot classification.
%We further enhance our network with a non-linear distance metric function, and outperforms other metric-based few-shot classification methods on the public miniImageNet benchmark.
We present a new efficient framework for few-shot learning, where the classifier must recognize new visual concepts with just a few labeled samples.
Our method aims to meta-learn a transferable feature embedding via a deep K-tuplet Network incorporated with learnable similarity to conduct few-shot classification.
Our network utilizes the relationship among the input samples for learning the generalized feature embedding not only on the training samples but also on the novel categories to alleviate the overfitting.
The key idea is to jointly compare multiple negative samples in the feature embedding learning.
Once trained, our network is able to extract discriminative features for unseen novel categories to facilitate the classification.
% motivations
%The network is trained on the miniImageNet training data and test on
Our approach sets the new state-of-the-art on the public miniImageNet benchmark, and also effectively generalizes for unseen novel class samples, even across different datasets.
\fi 
\end{abstract}

\begin{keyword}
Few-shot learning, metric learning, feature representation, deep learning.

\end{keyword}

\end{frontmatter}

\newcommand{\reviseagain}[1]{{\color{black}{#1}}}
\newcommand{\revise}[1]{{\color{black}{#1}}}
\def\ie{\emph{i.e.}}
\def\eg{\emph{e.g.}}
\def\etal{{\em et al.}}
\def\etc{{\em etc.}}
\renewcommand{\algorithmicrequire}{\textbf{Input:}}
\newcommand{\para}[1]{\vspace{.05in}\noindent\textbf{#1}}
\newcommand{\TODO}[1]{{\color{black}{#1}}}
\newcommand{\green}[1]{{\color{black}{#1}}}
\newcommand{\red}[1]{{\color{black}{#1}}}
\newcommand{\blue}[1]{{\color{black}{#1}}}
\newcommand{\lequ}[1]{{\color{magenta}{[LQ:#1]}}}
\newcommand{\xmli}[1]{{\color{black}{[Revise:#1]}}}

%\maketitle


%\input{abstract}
\section{Introduction}
%\begin{figure}
%	\centering
%	\includegraphics[width=1.0\linewidth]{figs/intro.pdf}
%	\caption{Our proposed framework for few-shot learning. (a) The training dataset are trained to learn feature embedding through triplet-like loss. (b) The well-learned training embedding are fed into similarity learning module to learn similarity among the query image and samples in the sample set. (c) The final prediction is the maximum similarity score.}
%	\label{fig:intro}
%\end{figure}
%%%%%%%%%%%%%%%%%%%%%%%%%%%%%%%%%%%%%%%%%%%%%%%%%%%%%
%\IEEEPARstart{L}{earning}

Learning from a few data is a hallmark of human intelligence, however, it remains a challenge for modern deep learning systems.
Recently, there has been a growing interest in few-shot learning~\cite{Czhang2019,snell2017prototypical,finn2017model,Sachin2017,wertheimer2019few,li2019revisiting,chen2019closer,bertinetto2018metalearning,jamal2018task,li2019ctm,lee2019meta,kim2019edge,liu2019large,lifchitz2019dense,liu2018learning,sun2019meta,cheny2019multi,chen2019imageblock,chen2019imagedeform,long2018object,schwartz2018delta,oreshkin2018tadam,li2019large,liu2019lcc,zhang2016contextual,jamal2019task}, which aims to recognize new visual concepts with just a small amount of labeled data for training.
In other words, the goal of few-shot learning is to classify unseen data instances (query examples) into a set of new categories, given just a small number of
labeled instances in each class (support examples).
%The ability to learn from a few examples is a hallmark
%of human intelligence, yet it remains a challenge for modern machine learning systems. This problem has received
%significant attention from the machine learning community
%recently where few-shot learning is cast as a meta-learning
%problem (e.g., [22, 8, 33, 28]).
%% Definition
%\IEEEPARstart{D}espite the successes in many computer vision problems such as object detection~\cite{ren2015faster}, semantic segmentation~\cite{long2015fully}, and object classification~\cite{he2016deep}, existing deep learning methods often struggle with the challenge of having limited labeled data.
%
%Hence, there has been a recent growing interest in few-shot learning, which aims to recognize new visual concepts with just a small amount of labeled data for training.
%
In this work, we focus on the case of few-shot classification, where {\em only a few labeled examples per class\/} are given.

%Deep learning techniques have achieved remarkable successes in various computer vision problems such as object detection~\cite{ren2015faster}, semantic segmentation~\cite{long2015fully}, and object classification~\cite{he2016deep}.
%However, current deep learning techniques struggle with the problem of limited annotated data, as they often require a large number of labeled data as inputs.
%%
%For this reason, there has been a recent increasing interest in the few-shot learning, which focuses on designing learning algorithms that specifically allow for recognizing new visual concepts with small labeled training sets.
%Few-shot classification aims to recognize novel visual categories from very few labeled examples.
%Here, we focus on the case of few-shot classification, where the given classification problem is assumed to contain only a few labeled examples per class.

%%%%%%%%%%%%%%%%%%%%%%%%%%%%%%%%%%%%%%%%%%%%%%%%%%%%%

Obviously, naively fine-tuning a model on the novel labeled data would easily overfit the {\em few\/} given data.
Hence, data augmentation and regularization~\cite{srivastava2014dropout,lee2015deeply} are often employed to somehow relieve the overfitting.
Later, the meta-learning paradigm~\cite{Sachin2017,santoro2016meta,finn2017model,munkhdalai2017meta} shed light to the few-shot learning problem; several metric learning-based methods~\cite{koch2015siamese,vinyals2016matching,snell2017prototypical,sung2018learning} were developed.
% exploited for the few-shot classification task, where the main idea is to first learn contextual task-specific knowledge from existing training tasks and then to apply the knowledge to perform few-shot test tasks on novel classes.
For instance, the \emph{matching network}~\cite{vinyals2016matching} uses an end-to-end trainable $k$-nearest neighbors algorithm on the learned embedding of the few labeled examples ({\em support set}) to predict the classes of the unlabeled samples ({\em query set}), while
%Matching networks can be interpreted as a weighted nearest-neighbor classifier applied within an embedding space.
%
the \emph{prototypical network}~\cite{snell2017prototypical} further builds a pre-class prototype representation.
More recently, Sung et al. presented the \emph{relation network}~\cite{sung2018learning}, which learns a nonlinear distance metric via a shallow neural network instead of using a fixed linear distance metric,~eg, Cosine~\cite{vinyals2016matching} and Euclidean~\cite{snell2017prototypical}.
These methods used sampled mini-batches called \emph{episode} to train an end-to-end network, aiming at making the training process more faithful to the test environment.
Although these methods utilize deep networks to extract expressive deep features, they do not take full advantages of the relationship among the input samples.
% for enhancing the features.
%
%However, these methods still suffer from having limited generalization ability for unseen class samples.
%
%Hence, we are motivated to explore whether a power feature embedding can be transferable for unseen class samples and improve the generalization for few-shot new category classification?
Hence, we are motivated to explore strategies to improve the feature embedding in terms of their efficiency to be transferable to handle unseen class samples and their generality for few-shot classification.
Although the triplet-like feature embedding is a longstanding topic in the computer vision area, its importance and effectiveness for the few-shot classification is neglected by the community.
%
%From another perspective, various deep metric learning techniques are developed for applications such as image retrieval~\cite{hoffer2015deep}, face recognition~\cite{schroff2015facenet} and person re-identification~\cite{xiao2017joint} problems.

%% Our work
In this work, we revisit  metric learning and investigate the potential of triplet-like feature embedding learning for few-shot classification.
We aim to {\em meta-learn a feature embedding\/} that performs well, not only on the training classes but more importantly, on the novel classes.
%A particular challenge of few-shot classification is to alleviate the overfitting, while enabling a classification on the unseen classes with very few examples.
%The utility of metric learning is shown due to the generalization of the learned feature embedding.
%
Specifically, the feature embedding should map the similar samples close to one another and dissimilar ones far apart.
%
%This is well aligned with the philosophy of triplet-like learning, but the traditional triplet network only interacts with a single negative sample per update, while few-shot classification requires a comparison with multiple query samples, typically of different classes.
%As a result, the learned embedded feature with traditional triplet loss is unlikely faithful to the test environment.
\revise{
This is well-aligned with the philosophy of triplet-like learning.
However, the general triplet network only interacts with a single negative sample per update, while few-shot classification requires a comparison with multiple query samples, typically of different classes.
\revise{
Hence, we formulate an improved triplet-like metric learning, namely the \emph{deep $K$-tuplet Network}, to improve the few-shot classification.
Particularly, the \emph{deep $K$-tuplet Network} generalizes the triplet network to allow joint comparison with $K$ negative samples in each mini-batch.}
It makes the feature embedding learning process more faithful to the few-shot classification problem with improved feature generalization.
Moreover, we present the \emph{semi-hard mining} sampling technique, an effective sampling strategy to sample informative hard triplets.
}
%Instead of sampling multiple query images from the same class~\cite{vinyals2016matching}, we randomly sample one query (anchor) images from one class to reduce the correlation of samples in one mini-batch.
%
Hence, we can speed up the convergence and stabilize the training procedure.

%%%
Our technique is simple yet powerful, and can be seamlessly incorporated with the learnable non-linear distance metric~\cite{sung2018learning} for few-shot classification.
To demonstrate the generalization capability of our presented few-shot classification framework,
we train our model on the miniImageNet dataset~\cite{vinyals2016matching}, and conduct few-shot classification, not only on the miniImageNet testing data, but also on other novel classes in other datasets (e.g., Caltech-101, CUB-200, Stanford Dogs and Cars).
Experimental results demonstrate that our method effectively generalizes for unseen novel class samples, even across different datasets.
%Although the training and testing classes in miniImage are different, there are very similar, while the different classes on different datasets are more different.\phil{this sentence is hard to follow...}

%\xm{In this work, we present a very simple yet powerful method for few-shot classification. we revisit  metric learning and investigate the potential of triplet-like feature embedding learning for few-shot classification. We aim to {\em meta-learn a feature embedding\/} that performs well, not only on the training classes but more importantly also on the novel classes.

The main contributions of this work are threefold:
\begin{enumerate}	
	%
%	\item
%	We revisit metric learning and demonstrate that a good feature embedding can bring substantial improvements to few-shot classification.
%	%
%	\item
%	We present the \emph{deep K-tuplet Network} to effectively learn the discriminative feature embedding on unseen class samples for few-shot learning.
%	%
%	\item
%	Our approach outperforms  state-of-the-art metric learning based few-shot classification methods on the miniImageNet dataset and achieves a remarkable success when generalizing to novel categories on completely different datasets.
%	%

%	\item
%	We provide a consistent comparative analysis of several representative few-shot methods and also reveal that a good feature embedding can bring substantial improvements to few-shot classification.

	\item
	We present a simple and powerful baseline method to investigate the importance of feature embedding for few-shot classification, where the effectiveness of feature embedding is neglected by previous works.

	\item
	 We present the \emph{deep $K$-tuplet Network} to effectively learn the discriminative feature embedding on unseen class samples for few-shot learning.
	 %By extending the triplet-network to deep K-tuplet network, %we present a simple yet powerful baseline method for few-shot classification, which achieves competitive performance over other methods.
	 Our method outperforms other metric-based methods and achieves competitive performance over other meta-based methods on the miniImageNet.
	
	 \item More importantly, prior works evaluated the few-shot learning within one dataset,~\ie, the novel classes and base classes are sampled from the same dataset. This experiment setting may be not representative in the real world setting. We establish a new experimental setting for evaluating the cross-domain generalization ability for few-shot classification algorithms. Our result generalized on CUB-200, Stanford Dogs, Stanford Cars and Caltech-101 excels other methods, showing the excellent cross-domain generalization capacity of our method.
	
\end{enumerate}

\section{Related Work}
Few-shot learning is an important area of research.
Early works on the few-shot learning focused on generative models and inference strategies~\cite{fei2006one,lake2011one}. 
In~\cite{fei2006one}, the authors assumed that one can utilize knowledge coming from previously-learned classes to make predictions on new classes only with one or few labels.
However, these methods do not involve deep learning.
Recently, with the success of deep learning, significant progress has been achieved in the few-shot learning area. 
%%Methods can be roughly categorized as metric-based methods, optimization-based methods, and sequence based methods.

%\reviseagain{Here we compare our method with the recent few-shot learning~\cite{Duan2020, hu2017sharable,lu2017discriminative}}. 

\subsection{Meta-learners for Few-Shot Learning}
\revise{
One category of the few-shot learning is meta-learner based methods~\cite{santoro2016meta,finn2017model,Sachin2017,munkhdalai2017meta,rusu2018meta}.
The meta-learning algorithm (MAML)~\cite{finn2017model} used a model agnostic meta-learner to train a good basic model on a variety of training tasks, such that given a new task with only a few training samples, a small number of gradient steps is sufficient to produce a good generalization model. 
Ravi \& Larochelle~\cite{Sachin2017} further proposed an LSTM-based meta-learning model to learn the optimization algorithm of training a network, where the LSTM updates the weights of a classifier for a given episode. 
Both methods, however, need to fine-tune the basic model on the target problem.
Munkhdalai \& Yu~\cite{munkhdalai2017meta} introduced a novel meta-learning architecture that learns meta-level knowledge across tasks and produces a new model via fast parameterization for rapid generalization.
Santoro~\etal~\cite{santoro2016meta} introduced a memory-augmented neural network to quickly encode and retrieve new data and make accurate predictions with only a few samples.
Lately, some other works~\cite{mishra2018simple, gidaris2018dynamic,Act2Param} focused on meta-learners for few-shot classification.
However, all these methods need to fine-tune or update the parameters for new unseen tasks, while our method performs the target tasks based entirely on feed-forward without requiring further parameter updates.}

\begin{figure*}[t]
	\centering
	\includegraphics[width=1.0\linewidth]{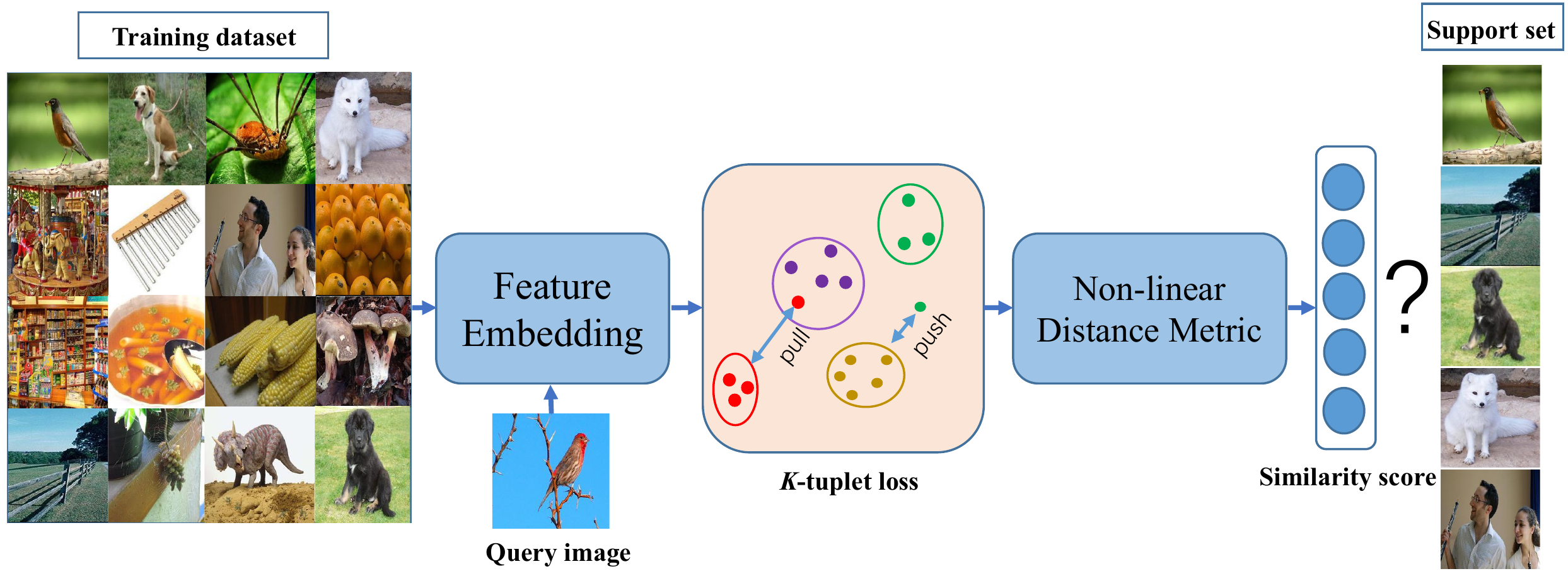}
	\caption{The framework of \emph{deep $K$-tuplet Network} for few-shot learning. We train a  embedding network to learn transferable feature embedding via the $K$-tuplet loss from the training dataset. The anchor interacts with multiple negative images in the tuplet, and contributes to the discriminative features. The well-learned embedding features are fed into the non-linear distance metric module to learn similarity among the query image and samples in the support set. Finally, we perform few-shot classification on the novel category.}
	\label{fig:framework}
\end{figure*}

\subsection{Deep Metric Learning}
\reviseagain{
Our work is related to deep metric learning, which involves a large volume of metric learning methods~\cite{bellet2013survey,hadsell2006dimensionality,hoffer2015deep,sohn2016improved,liu2018transductive}.
Below, we briefly review the more relevant ones.
The goal of metric learning is to minimize intra-class variations and maximize inter-class variations.
Early works used the siamese architecture~\cite{chopra2005learning,hadsell2006dimensionality} to capture the similarity between images.
The recent works~\cite{wang2014learning,schroff2015facenet,hoffer2015deep} adopted the deep networks as the feature embedding function and used triplet losses instead of pairwise constraints to learn the metric.
These metric learning strategies have been widely used in image retrieval~\cite{wang2014learning}, face recognition~\cite{schroff2015facenet,taigman2014deepface, lu2017discriminative,hu2017sharable} and person re-identification~\cite{xiao2017joint,hu2017sharable}.
For example, Lu~\etal~\cite{lu2017discriminative} proposed a discriminative deep metric learning method for face and kinship verification, where the distance of each positive pair is reduced and that of each negative pair is enlarged. 
Hu~\etal~\cite{hu2017sharable} proposed a multi-view metric
learning (MvML) to jointly learn an optimal combination of multiple distance metrics on multi-view representation. It learns a shared representation for different views and the method is applied on face verification, kinship verification, and person re-identification.
Duan~\etal~\cite{Duan2020} presented a deep adversarial metric learning
(DAML)to generate synthetic hard negatives
from the observed negative samples, where the potential hard negatives are generated to the learned metric as complements.
More recently, Wu~\etal~\cite{wu2018improving} presented a feature embedding method based on neighborhood component analysis.
These works show that combining deep models with proper objectives is effective in learning the similarities.
Unlike these methods, we consider using triplet-like networks to improve the feature discrimination on the unseen class images for few-shot learning problem.
}

\subsection{Metric Learning for Few-shot Learning}
The second branch are metric based approaches~\cite{snell2017prototypical,vinyals2016matching,sung2018learning,ren2018meta,garcia2017few,li2019lgm,koch2015siamese,wu2018improving}.
Metric learning based methods learn a set of project functions (embedding functions) and metrics to measure the similarity between the query and samples images and classify them in a feed-forward manner.
\revise{
The key difference among metric-learning-based methods lies in how they learn the metric.}
Koch~\etal~\cite{koch2015siamese} presented the siamese neural networks to compute the pair-wise distance between samples, and used the learned distance to solve the one-shot learning problem via a K-nearest neighbors classification.
\revise{
Vinyals~\etal~\cite{vinyals2016matching} designed end-to-end trainable k-nearest neighbors using the cosine distance on the learned embedding feature, namely matching network.}
%They also employed an attention mechanism to consider all the samples when computing the pair-wise distance between the query and support samples. 
%which uses an attention mechanism over a learned embedding of the labeled set of examples (the support set) to predict classes for the unlabeled points (the query set).
%Matching networks can be interpreted as a weighted nearest-neighbor classifier applied within an embedding space. 
%%
Lately, Snell~\etal~\cite{snell2017prototypical} extended the matching network by using the Euclidean distance instead of the cosine distance and building a prototype representation of each class for the few-shot learning scenario, namely prototypical network.
%The key idea is that the prototype is computed by taking the average of embedding vectors of samples from the same class
%%
Mehrotra \& Dukkipati~\cite{mehrotra2017generative} trained a deep residual network together with a generative model to approximate the expressive pair-wise similarity between samples.

Recently, Ren~\etal~\cite{ren18fewshotssl} extended the prototypical network to do semi-supervised few-shot classification, while Garcia~\etal~\cite{garcia2017few} defined a graph neural network to conduct semi-supervised and active learning. 
%in~\cite{garcia2017few}, a metric learning method based on graph neural networks has been proposed for few-shot classification, where it uses graph structure to model the relation between sample sets and query sets.
%%
Sung~\etal~\cite{sung2018learning} argued that the embedding space should be classified by a nonlinear classifier and designed the {\em relation module} to learn the distance between the embedded features of support images and query images. 
The relation network extends the matching network and prototypical network by including a learnable nonlinear comparator.
Notably, the prototypical networks~\cite{snell2017prototypical}, siamese networks~\cite{koch2015siamese}, and relation net~\cite{sung2018learning} all adopt the episode-based training strategy, where each episode is designed to mimic few-shot learning.
More recently, Li~\etal~\cite{li2019finding} proposed category traversal module (CTM) to look at all
categories in the support set to find task-relevant features.
Li~\etal~\cite{li2019revisiting} present the deep nearest
neighbor neural network to improve the final classification in the few-shot learning.
\revise{Although the excellent performance achieved in the few-shot classification, the importance of feature embedding has not paid sufficient attention. }

\if 1 
\subsection{Long-tail}
\cite{li2019large}
\cite{liu2019lcc}
\fi 

\section{Method}
\label{sec:approach}
\subsection{Overview}
\label{sec:fewshotlearning}
Few-shot classification involves three datasets: a training set $\mathcal{D}_{train}$, a support set $\mathcal{D}_{supp}$, and a query set $\mathcal{D}_{query}$.
In short, we want to train a model to learn transferable knowledge from $\mathcal{D}_{train}$, and apply the knowledge in the testing phase to classify the samples in $\mathcal{D}_{query}$ given $\mathcal{D}_{supp}$.
\begin{itemize}
\item
$\mathcal{D}_{train} = \left\{ (x_i, y_i) \right\}_{i=1}^{N}$ is used for training the model, where $x_i$ is a training image, $y_i \in \mathcal{C}_{train}$ is the label of $x_i$, and $N$ is the number of training examples.
\item
$\mathcal{D}_{supp}  = \left\{ (x_j, y_j) \right\}_{j=1}^{M}$ is the set of $M$ labeled examples given in the testing phase, where $y_j \in \mathcal{C}_{supp}$ is the label of $x_j$ but $\mathcal{C}_{train} \cap  \mathcal{C}_{supp} = \varnothing$.
\item
Given $\mathcal{D}_{query} = \left\{ x_j \right\}_{j=1}^{n}$, the goal of few-shot classification is to classify the samples in $\mathcal{D}_{query}$.

%\item 
%%Sample set/query set split is designed to mimic the support/test set in the inference setting, where 
%
%$D_{sample}$
%sample set is constructed by randomly select $C$ classes from the training set with $K$ labeled samples in each class.  
%The reminder of those $C$ class

%The episode is formed by sample $K$ labeled data from $D_{train}$ and query reminders of to serve as query. 

%
\end{itemize}
Note that the support set $\mathcal{D}_{supp}$ and the query set $\mathcal{D}_{query}$ share the same label space.
If the support set has $K$ labeled examples for each of the $C$ classes in $\mathcal{C}_{supp}$,~\ie, $M=C\times K$, then the few-shot problem is called $C$-way $K$-shot. 

Figure~\ref{fig:framework} overviews our few-shot learning framework.
First, we meta-learn a transferable feature embedding through the \emph{deep K-tuplet network} with the designed $K$-tuplet loss from the training dataset. 
The well-learned embedding features of the query image and samples in the support set are then fed into the non-linear distance metric to learn the similarity scores.
Further, we conduct few-shot classification based on these scores.
%%%%
\subsection{Meta-learn Feature Embedding}
\label{sec:featureembedding}
%\xm{our featrue embedding can achieve better results on CUBbird and Car dataset. So We can claim our feature embedding is a meta-learner, that is suitable for novel class. Since we randomly sample triplets, so we learn many mini-tasks in each update.}To learn transferable knowledge from $\mathcal{D}_{train}$ requires the learning of a nonlinear mapping that models the class relationship among the samples,i.e., mapping similar samples close to one another, while dissimilar ones far apart.
Such nonlinear mapping should be generalizable to work with samples of novel classes, meaning that the mapping should preserve the class relationship on the unseen class samples in $\mathcal{D}_{supp}$ and $\mathcal{D}_{query}$. 
We adopt a {\em triplet-like network\/} to learn the feature embedding on $\mathcal{D}_{train}$.

Specifically, for an input image $x_i$, function $f(\cdot;\theta):\mathcal{X}  \rightarrow \mathbb{R}^d$ maps $x_i$ to an embedding vector $f(x_i)$,
where $\theta$ denotes the parameters of the embedding function;
$d$ is the dimension of the embedded features, and
$f(x_i)$ is usually normalized to unit length for training stability and comparison simplicity. 
To learn parameter $\theta$, the traditional triplet loss is widely used, where the objective is based on a relative similarity or distance comparison metric on the sampled pairs.
%that can be consistent across different classes. 
%
In short, the training samples are randomly selected to form a triplet ($x_a, x_p, x_n$) with an anchor sample $x_a$, a positive sample $x_p$, and a negative sample $x_n$.
The label of the selected samples in a triplet should satisfy $y_a = y_p \neq y_{n}$.
The aim of the loss is to pull $f(x_a)$ and $f(x_p)$ close to each other, while pushing $f(x_a)$ and $f(x_n)$ far apart.

\if 0
Hence, the optimization objective is formulated as:
%
\begin{equation}
L(x_a, x_p, x_n) =  \left [ \left \| f_a - f_p \right \|^2 - \left \|  f_a - f_n\right \|^2 + \alpha \right ]_+,
\end{equation}
where $\left [ \cdot \right ]_+ = \max(0, \cdot)$ denotes the hinge loss function and $\alpha$ is the margin that pushes the distance of negative pairs to be larger than the distance of positive pairs.
We omit $x$ from $f(x)$ for simplicity.
\fi

\revise{
However, the above traditional triplet loss interacts with only one negative sample (and equivalently one negative class) for each update in the network, while we actually need to compare the query image with multiple different classes in few-shot classification.
Hence, the triplet loss may not be effective for the feature embedding learning, particularly when we have several classes to handle in the few-shot classification setting. 
Inspired by~\cite{sohn2016improved}, we generalize the traditional triplet loss to a tuplet loss with $K$-negatives, namely $K$-tuplet loss, to allow simultaneous comparison jointly with $K$ negative samples, instead of just one negative sample, in one mini-batch.
This extension makes the feature comparison more effective and faithful to the few-shot learning procedure, since each update, the network can compare a sample with multiple negative classes altogether.}

\begin{figure*}[!t]
	\centering
	\includegraphics[width=1.0\linewidth]{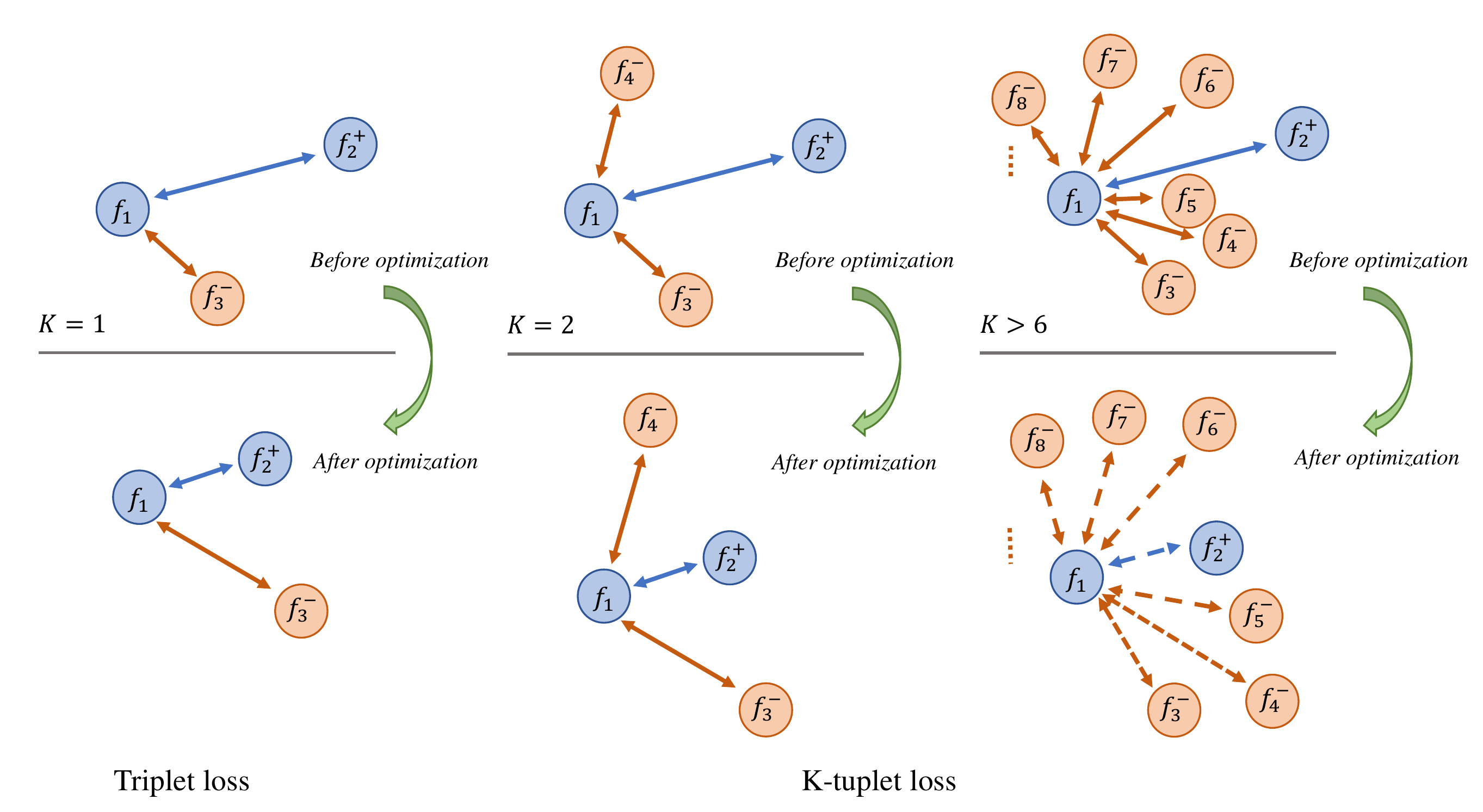}
	\caption{\revise{The visualization of $K$-tuplet loss function. When $K$ equals to 1, $K$-tuplet loss becomes triplet loss.}}
	\label{fig:k-tuplet}
\end{figure*}

\revise{
In particular, we randomly choose the $K$ negative samples $x_{n_i}, i=\{1,2,.., K\}$ to form into a triplet.
%\phil{I remembered XM said that she has tried some strategies here, e.g., finding some negative samples that are close to the anchor sample}
Accordingly, the optimization objective is formulated as:
\begin{equation}
\mathcal{L}(x_a, x_p, x_{n_i}) = \frac{1}{K}\sum_{i \in U} {\left [ \left \| f_a - f_p \right \|^2 - \left \|  f_a - f_{n_i}\right \|^2 + \alpha \right ]_+}, K=|U|
\label{eq1}
\end{equation}
where $\left [ \cdot \right ]_+ = \max(0, \cdot)$ denotes the hinge loss function, $\alpha$ is the hyperparameter margin and $U$ denotes the set of triplets, and we write $f(x)$ as $f$ to omit $x$ for simplicity.}
For the anchor sample $x_a$, the optimization shall maximize the distance to the negative samples $x_{n_i}$ to be larger than the distance to the positive sample $x_p$ in the feature space.
To form one mini-batch to train the network, we randomly select $B$ anchor samples from the training set, where $B$ is batch size.
For each anchor sample $x_a$, we then randomly select another positive sample $x_p$ of the same class as $x_a$ and further randomly select $K$ other negative samples whose classes are different from $x_a$.
Among the $K$ negative samples, their class labels may be different.~\revise{
Figure~\ref{fig:k-tuplet} visualizes the $K$-tuplet loss and triplet loss.
When $K$ equals to 1, $K$-tuplet loss becomes triplet loss. 
The classification accuracy is improved with a larger $K$,since the anchor sample interacts with more samples in one mini-batch and makes the gradient more stable.  However, when $K$ increases, the computational burden increases and it becomes too heavy-lifting to perform standard optimization, i.e., stochastic gradient descent (SGD) with a mini-batch. 
In other words, the classification accuracy would decrease as K continues to increase.
%In other words, the classification accuracy would be decreased with a bigger $K$. 
From our experiments, we can achieve the best performance when setting $K$ to 5. 
%Compared with the traditional triplet loss, each forward update in our $K$-tuplet loss considers more inter-class variations, thus making the learned feature embedding more discriminative for samples from different classes.
Compared with the traditional triplet loss, more inter-class variations have been considered in each forward update by using our $K$-tuplet loss, thus making the learned feature embedding more discriminative for samples from different classes. 
}

\if 0
Most few-shot classification methods employ sampled mini-batches, namely \emph{episode}, to mimic the test scenario during the training.
%
In our work, we explore another sampling strategy to better learn the above feature embedding function.
%
Specifically, we randomly select $B$ anchor samples from the training set.
For each anchor sample $x_a$, we then randomly select another positive sample $x_p$ of the same class with $x_a$ and further randomly select $K$ other negative samples whose class labels are different from $x_a$.
Note that these $K$ negative samples may have different class labels.
%
Compared with the standard episode-based sampling strategy, our sampling strategy can contribute to create a general feature embedding among the given classes.
%increase the data diversity i.e., the different class samples) given the fixed number of images in the mini-batch.

\fi

%\begin{figure}
%	\includegraphics[width=1.0\linewidth]{figs/method.pdf}
%	\caption{Illustration of tuplet with K-negatives.}
%	\label{fig:tupletloss}
%\end{figure}

%%%%%%%%%%%%%%%%%%%%%%%%%%%%%%%%%%%%%%%%%%%%%%%%%%%%%%%%

\subsection{Efficient Training with Semi-hard Mining}

\revise{The semi-hard mining strategy is motivated by the observation that when the model starts to converge, the “well-learned easy samples” will obey the margin and could not contribute to the optimization in the learning process. However, the ``hard samples'' still fail to satisfy the optimization goal. This phenomenon degrades the model performance and also slows down the convergence of the training. We, thereby, design a semi-hard mining strategy to sample more informative hard triplets in each mini-batch when the model starts to converge. The informative hard triplets are selected by whether the condition in the loss function is satisfied or not. The loss function of semi-hard mining can be described as the following:
\begin{equation}
\begin{split}
\mathcal{L}_{semi-hard}(x_a, x_p, x_{n_i}) = \frac{1}{s}\sum_{i \in S} {\left [ \left \| f_a - f_p \right \|^2 - \left \|  f_a - f_{n_i}\right \|^2 + \alpha \right ]_+}, \\
{\rm where} \ S=\{i\in U \mid \|f_a-f_{n_i}\|^2-\|f_a-f_p\|^2\geq\alpha\} \ {\rm and} \ s=|S|
\end{split}
\label{eq2}
\end{equation}
where $\left [ \cdot \right ]_{+}  = \rm{max} (0, \cdot)$ denotes the hinge loss function and $\alpha$ is hyperparameter margin. 
$x_a, x_p, x_n$ denote an anchor, positive and negative sample, respectively. $s$ is the number of elements in set $S$, where set $S$ represents the triplets that are selected as informative and hard. We write $f(x_a)$ as $f_a$ to omit $x$ for simplicity.  

This semi-hard loss function is utilized when the model starts to converge (80 epochs in our experiments) and we continue to fine tune it for 100 epochs. We utilize Adam optimizer with a learning rate of 0.001 to train the network.
We analyze the effectiveness of this technique in Table~\ref{tab:semihard}. 
From our experiments, we can see that this semi-hard mining strategy helps improve the training efficiency and contributes to the learning of feature embedding. 
}

\if 1
\revise{
When training with the $K$-tuplet loss, the individual update of one mini-batch may be unstable. This is because when the model starts to converge, the well-learned samples obey the margin and can not contribute to the gradients in the learning process.
This phenomenon degrades the model capacity and slows down the convergence of the training.  
We, thereby, design a semi-hard mining strategy to sample more informative hard triplets in each mini-batch when the model starts to converge.
Specifically, we first randomly sample $B$ triplets according to the strategy in Section~\ref{sec:featureembedding}. 
Then, we intentionally check if the sampled triplets obey the margin or not.
We remove those well-learned triplets that have already satisfied the margin, and sample remainder triplets from the training sets to form a mini-batch.
The final objective is then calculated on the new mini-batch. 

\begin{equation}
\mathcal{L}(x_a, x_p, x_{n_i}) = \frac{1}{K}\sum_{i=1}^{K} {\left [ \left \| f_a - f_p \right \|^2 - \left \|  f_a - f_{n_i}\right \|^2 + \alpha \right ]_+},
\end{equation}

From our experiments, we can see that this \emph{semi-hard mining} strategy helps improve the training efficiency and contributes to the learning of feature embedding.  
}
\fi

\subsection{Non-linear Distance Metric Learning}
Furthermore, we adopt the non-linear distance metric module~\cite{sung2018learning} to learn to compare the embedded features in few-shot classification.
%during the \emph{episode} training to further increase \phil{improve?} the feature comparison \phil{quality?}.
%
Given image $x_s$ from the support set and image $x_q$ from the query set,
their similarity score is learned by concatenating $f_\theta(x_q)$ and $f_\theta(x_s)$ and then feeding the combined feature into a non-linear distance metric.
The non-linear distance metric has two convolutional blocks and two fully-connected layers.
Each convolutional block consists of a $3\times3$ convolution with 64 channels followed by a batch normalization, an ReLU activation function, and a $2\times2$ max-pooling.
The fully-connected layers have 8 and 1 outputs, followed by a sigmoid function to get the final similarity scores between the query image $x_q$ and samples in the support set.
In the end, our non-linear distance metric learns to produce the similarity score by calculating the mean square error loss, following the same spirit as~\cite{sung2018learning}.

\begin{figure}[htbp]
	\centering
	\includegraphics[width=0.9\linewidth]{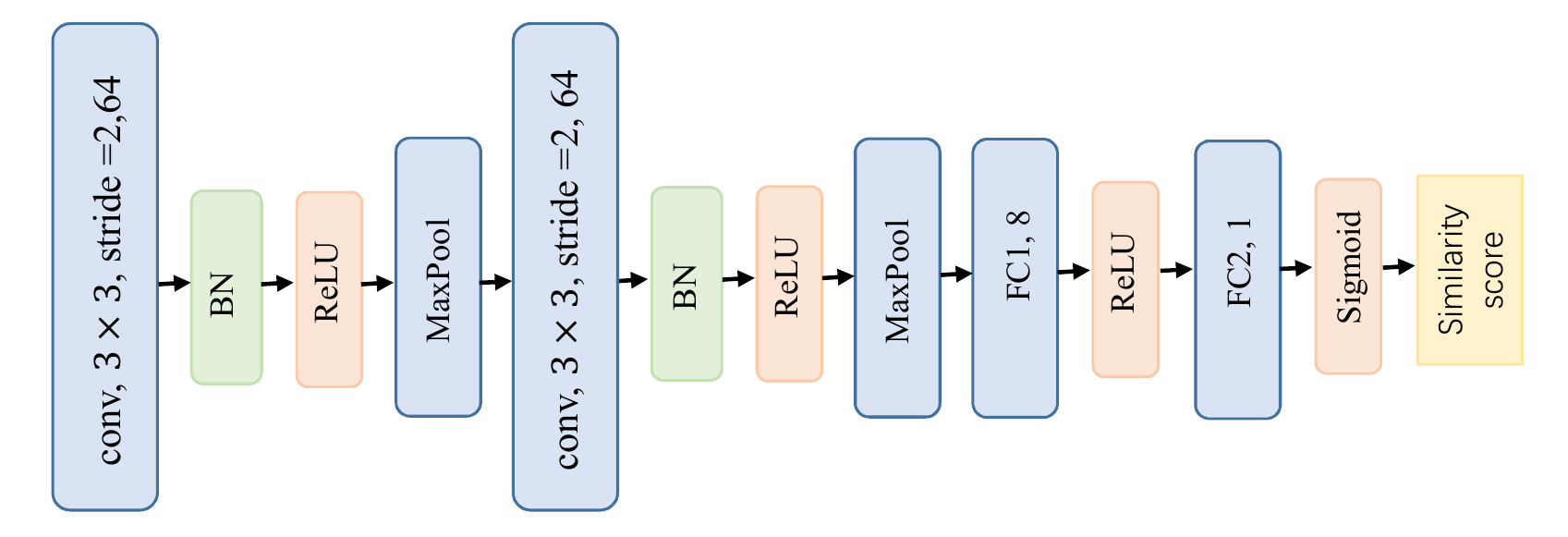}
	\caption{The detailed network architecture of our nonlinear metric learning module. The last number in each box denotes the number of feature channels.}
	\label{non-linearmetric}
\end{figure}

%\subsection{Architectures of Nonlinear Metric Learning Module}
Figure~\ref{non-linearmetric} shows the detailed network architecture of our nonlinear metric learning module. 
The input is the concatenation of features from the images of the support set and the query set. 
The output is the similarity scores of the query images with images in the support set. 
The few-shot classification prediction is the label of the image that has the maximum similarity score in the support set. 

\subsection{Technique Details}
We employed the ResNet34 architecture~\cite{he2016deep} for learning the feature embedding.
%We further normalize the feature into unit length to avoid degeneracy.
When meta-learning the transferable feature embedding, we used Adam optimizer~\cite{kingma2014adam} with a learning rate of 0.001 and a decay for every 40 epochs.
We totally trained 100 epochs and adopted the semi-hard mining strategy when the loss starts to converge (at around 80 epoches).
To learn the non-linear distance metric, we followed the episode-based strategy and also employed the Adam optimizer with a learning rate of 0.001.
Different from the general episode sampling procedure, we sampled multiple episodes to form each mini-batch to train the non-linear distance metric.
This strategy increases the data diversity,~\ie, the number of different class samples) and makes the training more stable.

%\vspace*{-3mm}
%\paragraph{Evaluation procedure.}
We evaluate the accuracy of few-shot classification by averaging the  randomly-generated episodes from the training set, following~\cite{snell2017prototypical}. 
For 5-way 1-shot test, each query image is compared with five samples in the support set.
The prediction is the label of the sample that has the maximum similarity score within the support set. 
For 5-way 5-shot test, we sum the features of all the samples in each class in the support set as the feature map of the class and then follow the same procedure with 5-way 1-shot setting to get the query image label.

%In each iteration, we acquire the relation score i.e., similarity) between the embedding of test images with the embedding of images from the sample set via the relation module. 

%In 1-shot setting, we directly input the feature of the support image, while we element-wise sum over the feature embedding of all samples from each class to form this class' feature, following~\cite{sung2018learning}.
%each sample image represent a prototypical representation. In 5-shot setting, the prototypical representation is the averages of 5 features in each class.

%After learning a feature embedding function f that is universal across classes, one-shot prediction is straight forward. 
%For any test example xtest from the novel classes, we predict its label by finding its nearest neighbour from the one-shot instances in the embedding space and assigning it to the corresponding class If a probability distribution is needed as output, one can simply apply a softmax function:

\section{Experiments}
\label{sec:experiment}
We first evaluate our few-shot classification method on the public miniImageNet dataset.
We then show the generalization of our approach by directly evaluating on completely different datasets using the model trained with miniImageNet.
Lastly, we extensively analyze the different components of our method.

\subsection{Few-shot Classification on the miniImageNet}
%% dataset
The miniImageNet dataset is derived from the ILSVRC-12 dataset~\cite{russakovsky2015imagenet}, consisting of 60,000 color images with 100 classes and 600 samples per class.
In order to directly compare with state-of-the-art algorithms, we follow the splits introduced by Ravi and Larochelle~\cite{Sachin2017}, with 64, 16 and 20 classes for training, validation and testing, respectively. The validation dataset is used for monitoring generalization performance of the network only and not used for training the network.

We compare our approaches with several state-of-the-art methods reported on the miniImageNet~\cite{vinyals2016matching,snell2017prototypical,sung2018learning}, as shown in Table~\ref{tab:finaltable}.
Most of the existing methods employed the shallow neural network,~\ie, four convolutional layers, to extract the feature. Since our method is based on the well-learned feature embedding, the shallow embedding network did not make adequate usage of our method’s expressive capacity. Thus, we follow the recent works~\cite{mishra2018simple,lee2019meta,munkhdalai2017rapid} to use a deeper embedding network, \ie, ResNet, to prevent the underfitting.

Compared with metric-based methods, we can see that our method achieves the highest accuracy on 5-way 1-shot setting and very competitive accuracy on 5-way 5-shot setting, as shown in Table~\ref{tab:finaltable}.
Note that Li \etal~\cite{li2019revisiting} achieves 54.37 $\pm$ 0.36 \% and 74.44 $\pm$ 0.29 \% with ResNet backbone on 5-way 1-shot and 5-way 5-shot respectively.
However, our result outperforms their method on 5-way 1-shot and shows competitive result on 5-way 5-shot setting.  
We report the few-shot classification accuracy of our method using the $K$-NN classifier with the Euclidean distance on the embedded feature; see \emph{Ours+Euclid} in Table~\ref{tab:finaltable}.
In this setting, we remove the non-linear metric and use $K$ nearest neighbors ($K=1$) on the embedded features of query images and support images for classification.
It is observed that the Euclid version of our method still achieves the competitive results, showing the generalization and discrimination of the learned feature embedding on unseen novel categories.

\reviseagain{Beside metric-based methods, there are several state-of-the-art meta-learning approaches for the few-shot learning problem~\cite{rusu2018meta,Act2Param,gidaris2018dynamic,lee2019meta}. For example, Gidaris~\etal~\cite{gidaris2018dynamic} propose a dynamic-net, and report $ 56.20 \pm 0.86 $ (\%) on 5-way 1-shot and $72.81 \pm 0.62 $ (\%) on 5-way 5-shot setting.
Andrei A~\etal~\cite{rusu2018meta} proposed a latent embedding optimization (LEO) meta-learning approach that decouples the gradient-based adaptation procedure from the underlying high-dimensional space of model parameters. Their method achieved 61.76 $\pm$ 0.08 (\%) on 5-way 1-shot settings. 
More recently, Kwonjoon~\etal~\cite{lee2019meta} explored two properties of linear classifiers in meta-learning, \ie, implicit differentiation of the optimality
conditions of the convex problem and the dual formulation
of the optimization problem. Their method achieved  $62.64 \pm 0.61$ (\%) on 5-way 1-shot and 78.63 $\pm$ 0.46 (\%)  on 5-way 5-shot setting. However, this method is based on the ResNet 12 backbone and the direct comparison is not fair. As our work learns deep metrics in the embedding space, we mainly compare with metric-based approaches. More importantly, our method has only one single unified network, which is much simpler than these meta-learning-based methods with additional complicated memory-addressing architectures.
}

\begin{figure*}
	\centering
	\includegraphics[width=1\linewidth]{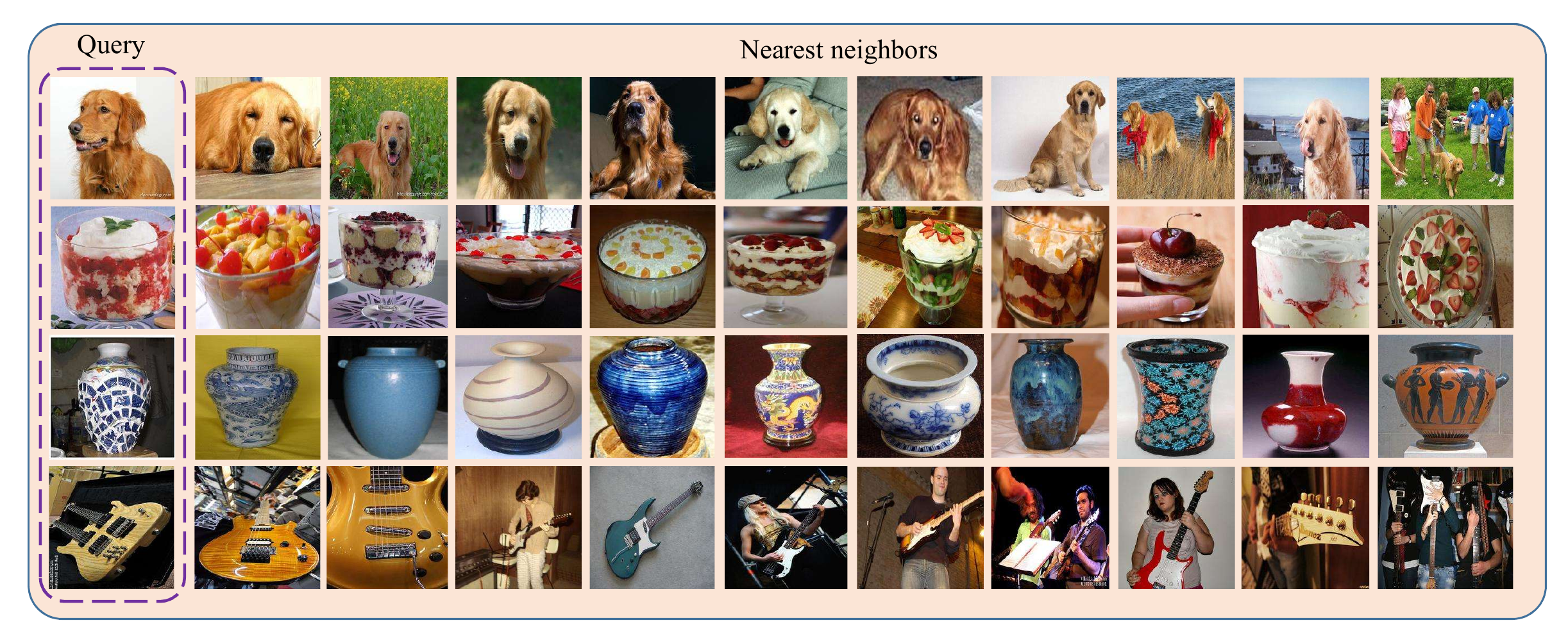}
	\caption{Nearest neighbors from the learned feature embedding of our method  on the miniImageNet testing dataset. Given a query image, we shows 10 nearest neighbor images.}
	\label{fig:nearest}
\end{figure*}

In Figure~\ref{fig:nearest}, we show the 10 nearest neighbor images of the query image on the miniImageNet testing dataset with the Euclid distance of our learned embedding features.
We can see our feature embedding preserves apparent visual similarity better and facilitates the accurate recognition.

\begin{table*}[h]
	\centering
	\caption{
		Average few-shot classification accuracies (\%) on the miniImageNet. Note that `-' denotes not reported. All accuracy results are averaged over 600 test eposides and are reported with 95\% confidence intervals. }
	\label{tab:finaltable}
	{\centering	
		\resizebox{0.7\textwidth}{!}{\begin{tabular}{c|c|c|c}
				\hline
				\toprule
				\multirow{2}{*}{Model}	   & \multirow{2}{*}{Year} & \multicolumn{2}{c}{ 5-way Acc. } \tabularnewline
				\cline{3-4} &
				& 	  1-shot &   5-shot \tabularnewline
				\hline
				Matching Nets~ \cite{vinyals2016matching} & 2016 NIPS  &  46.6 $\pm$ 0.8 & 60.0 $\pm$ 0.7
				\tabularnewline
				Meta-Learn LSTM \cite{Sachin2017} & 2017 ICLR  &   43.44 $\pm$ 0.77 & 60.60 $\pm$ 0.71
				\tabularnewline

				MAML \cite{finn2017model}  & 2017 ICML  &  48.70 $\pm$ 1.84 & 63.11 $\pm$ 0.92
				\tabularnewline	
				Meta Nets \cite{munkhdalai2017meta}  & 2017 ICML  &    49.21 $\pm$ 0.96 &  -
				\tabularnewline
				
				Proto Net~\cite{snell2017prototypical} & 2017 NIPS & 49.42 $\pm$ 0.78 & 68.20 $\pm$ 0.66
				\tabularnewline
				
				Proto Net (\textit{ResNet})~\cite{snell2017prototypical} & 2017 NIPS &	51.15 $\pm$ 0.85 &  69.02 $\pm$ 0.75 \tabularnewline
				
				Triplet ranking~\cite{ye2018deep} & 2018 Arxiv  & 48.76  & -  \tabularnewline
				
				GNN ~\cite{garcia2017few} & 2018  ICLR  &  50.33 $\pm$ 0.36 & 66.41 $\pm$ 0.63  \tabularnewline
				
				Masked Soft k-Means~\cite{ren18fewshotssl} & 2018 ICLR   & 50.41 $\pm$ 0.31  & 64.39 $\pm$ 0.24 \tabularnewline
				
				Relation Net~\citep{sung2018learning}  & 2018 CVPR   &   50.44 $\pm$ 0.82  & 65.32 $\pm$ 0.70
				\tabularnewline

				Relation Net (\textit{ResNet})~\citep{sung2018learning} & 2018 CVPR & 	52.13 $\pm$ 0.82 & 64.72 $\pm$ 0.72 \tabularnewline 			
				
				large margin few-shot~\cite{wang2018large} & 2018 Arxiv & 51.08 $\pm$ 0.69 &  67.57 $\pm$ 0.66		\tabularnewline

				SNAIL~\cite{mishra2018simple} & 2018 ICLR & 55.71 $\pm$ 0.99 & 68.88 $\pm$ 0.92 \tabularnewline

				%Dynamic-Net~\cite{gidaris2018dynamic}  & 2018 CVPR  & 56.20 $\pm$ 0.86  & \green{\textbf{73.00 $\pm$ 0.64}}    \tabularnewline
				%AdaResNet~\cite{munkhdalai2017rapid} & 2018 ICML  &  \green{\textbf{56.88 $\pm$ 0.62}}  & 71.94 $\pm$ 0.57
				%\tabularnewline
				
				R2D2~\cite{bertinetto2018metalearning} & 2019 ICLR & 51.2 $\pm$ 0.6 & 68.8 $\pm$ 0.1 \tabularnewline
				
				DN4~\cite{li2019revisiting} & 2019 CVPR   & 51.24 $\pm$ 0.74  & 71.02 $\pm$ 0.64 \tabularnewline
				
				%	DN4 (\textit{ResNet})~\cite{li2019revisiting} & 2019  CVPR  & 54.37 $\pm$ 0.36  & \TODO{\textbf{74.44 $\pm$ 0.29}} \tabularnewline
				%			DEML+Meta-SGD~\cite{zhou2018deep}$\dagger$  & 2018 arxiv & Meta & 58.49 $\pm$ 0.91 & 71.28 $\pm$ 0.69 \tabularnewline
				
				%	Transductive Prop Nets~\cite{liu2018transductive}  & 2019  &  55.51 $\pm$ 0.86 & 69.86 $\pm$ 0.65
				%	\tabularnewline
				\hline
				Ours+Euclid  & -  &  \blue{54.46 $\pm$ 0.89}  & 68.15 $\pm$ 0.65 \tabularnewline				
				
				\textbf{Ours} & -  & \TODO{\textbf{58.30 $\pm$ 0.84}}  & \blue{\textbf{72.37 $\pm$ 0.63}}  \tabularnewline

%		 \multicolumn{4}{c}{To take a whole picture of the-state-of-art methods}				
%		 \tabularnewline \hline 
%	
%
%		Dynamic-Net~\cite{gidaris2018dynamic}$\dagger$	& 2018 CVPR & 56.20 $\pm$ 0.86 &  72.81 $\pm$ 0.62 \tabularnewline   
%		 	
%			Activation to Parameter ~\cite{Act2Param} $\dagger$ & 2018 CVPR & 59.60 $\pm$ 0.41 & 73.74 $\pm$ 0.19
%			\tabularnewline
%			LEO~\cite{rusu2018meta}  $\dagger$ & 2019 ICLR  & 61.76 $\pm$ 0.08 & 77.59 $\pm$ 0.12   \tabularnewline
				\bottomrule
				\hline
			\end{tabular}	
		}		}
	\end{table*}
		
\begin{table*}[t]
	\centering
	\caption{Average few-shot classification accuracies (\%) on other datasets using the models trained with the miniImageNet. Note that all the experiments are conducted with the same network for fair comparison.}\label{tab:caltech101}

	\resizebox{0.9\textwidth}{!}{\begin{tabular}{cccccc}
		\hline
		\toprule
		\multicolumn{2}{c}{Dataset}  & {Proto Net~\cite{snell2017prototypical}} & {Relation Net~\cite{sung2018learning}} &  Cosface embed~\cite{wang2018cosface} &  {Ours}\tabularnewline
		\hline
		\multirow{2}{*}{Caltech-101} & {5-way 1-shot} & 53.28 $\pm$ 0.78 & 53.50 $\pm$ 0.88 & 57.22 $\pm$ 0.85 & \textbf{61.00 $\pm$ 0.81}  \tabularnewline
		& {5-way 5-shot} & 72.96 $\pm$ 0.67 & 70.00 $\pm$ 0.68 & 75.34 $\pm$ 0.69  & \textbf{75.60 $\pm$ 0.66}  \tabularnewline
		\hline
		\multirow{2}{*}{CUB-200} & {5-way 1-shot} & 39.39 $\pm$ 0.68 & 39.30 $\pm$ 0.66 & 39.60 $\pm$ 0.70  &\textbf{ 40.16 $\pm$ 0.68}    \tabularnewline
		& {5-way 5-shot} & 56.06 $\pm$ 0.66 & 53.44 $\pm$ 0.64 & 55.70 $\pm$ 0.66  & \textbf{ 56.96 $\pm$ 0.65} \tabularnewline
		\hline
		\multirow{2}{*}{Stanford Dogs} & {5-way 1-shot} & 33.11 $\pm$ 0.64 &  31.59 $\pm$ 0.65 &  43.16 $\pm$ 0.84 & \textbf{37.33 $\pm$ 0.65} \tabularnewline
		& {5-way 5-shot} & 45.94 $\pm$ 0.65 & 41.95 $\pm$ 0.62 & 49.32 $\pm$ 0.77 & \textbf{49.97 $\pm$ 0.66 }\tabularnewline
		\hline
		\multirow{2}{*}{Stanford Cars} & {5-way 1-shot} & 29.10 $\pm$ 0.75 & 28.46 $\pm$ 0.56 & 29.57 $\pm$ 0.70 &  \textbf{31.20 $\pm$ 0.58} \tabularnewline
		& {5-way 5-shot} & 38.12 $\pm$ 0.60 & 39.88 $\pm$ 0.63 & 40.78 $\pm$ 0.68 &
		\textbf{47.10 $\pm$ 0.62} \tabularnewline	
		\bottomrule
		\hline
	\end{tabular}}
\end{table*}

\if 1
To take a whole picture of the few-shot learning area, we
also report the results of the state-of-the-art meta-learning
based methods. We can see that our method is still competitive with these methods. Especially in the 5-way 1-shot setting, our method achieves 13.2\%, 9.6\%, 9.1\% and 2.1\% improvements over SNAIL, MAML, Meta Nets and Dynamic-Net, respectively.
The Dynamic-Net is a two-stage model, which pre-trains with all classes together before conducting the few-shot training,
while our method does not.
More importantly, our method is a single unified network and performs the inference in a feed-forward manner, which is much simpler than these meta-learning based methods that need additional complicated memory-addressing architectures or additional adaptation steps.
\fi
%\begin{figure}
%	\includegraphics[width=1.0\linewidth]{figs/backupfigs/accuracy.pdf}
%	\caption{Test accuracy on 5-way 1-shot and 5-way 5-shot in comparison with Relation Net~\cite{sung2018learning}, respectively. The implementation used the same network architecture. The figures shows that compared to relation net, our feature embedding can contribute to high performance and fast training.}
%	\label{unseen embedding}
%\end{figure}

\begin{figure*}
\centering
\includegraphics[width=1\linewidth]{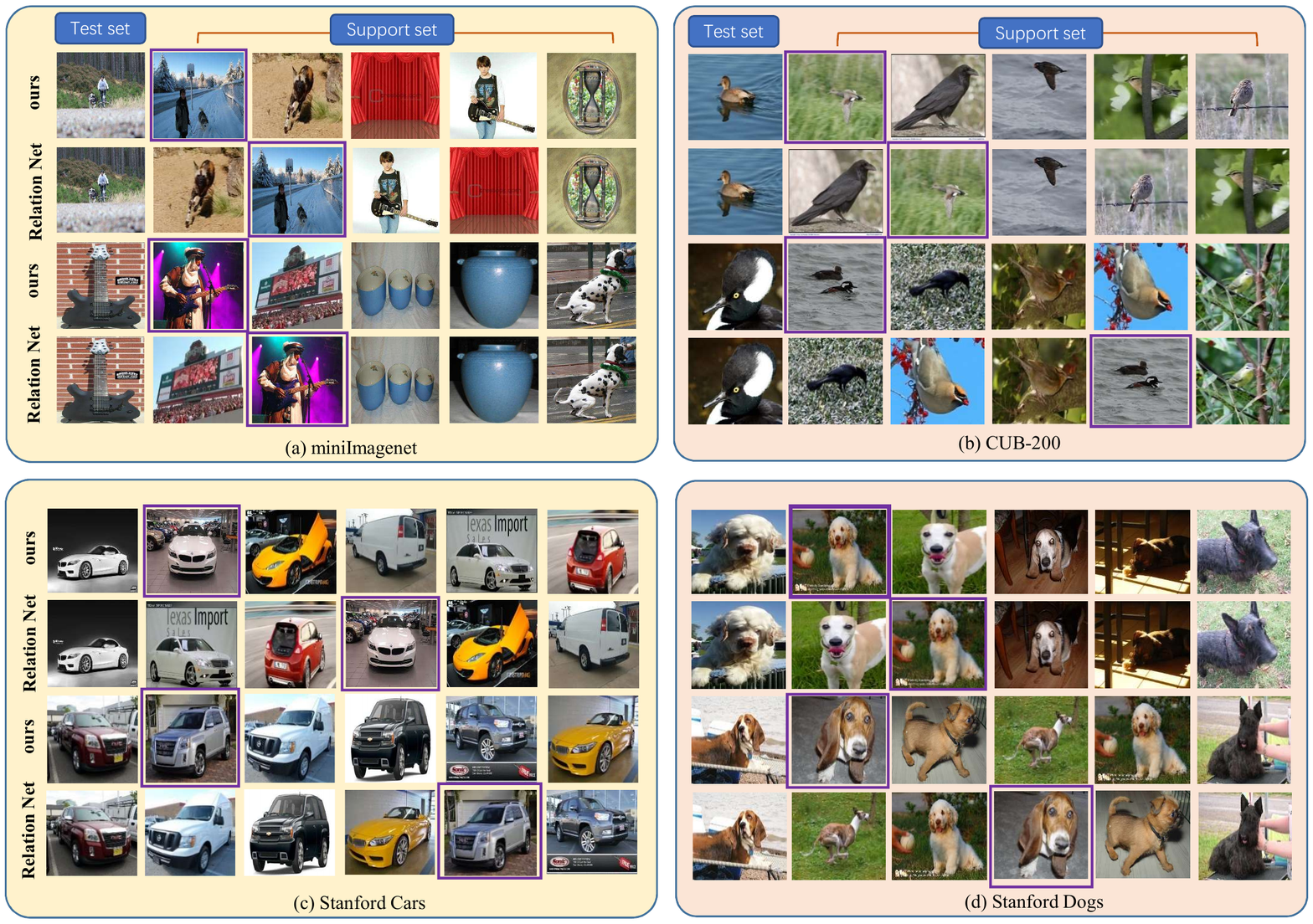}
\caption{Examples of visualized results of few-shot classification on (a) miniImageNet, (b) CUB-200, (c) Stanford Cars and (d) Stanford Dogs dataset. The images in the support set is sorted by the similarity with the test image (from left to right and only showing top-5 images). Purple box denotes the ground-truth class in the support set.}
\label{fig:cub200}
\end{figure*}

\if 1
\begin{table*}[t]
	\centering
	\caption{Generalization of models for few-shot classification on other dataset.}
	\label{tab:caltech101}{		\resizebox{2\columnwidth}{!}{	
	\begin{tabular}{ccccccccc}
		\toprule
		\multirow{2}{*}{Dataset}&  \multicolumn{2}{c}{Caltech-101}& \multicolumn{2}{c}{CUB-200} & \multicolumn{2}{c}{Standford Dogs} & \multicolumn{2}{c}{Standford Cars} \tabularnewline
		
		&   5way 1shot & 5way 5 shot & 5way 1shot & 5way 5shot & 5way 1shot & 5way 5shot & 5 way 1 shot & 5 way 5 shot \tabularnewline
		\hline
		Relation Net~\cite{sung2018learning} & 53.50 $\pm$ 0.88\%  & 70.00 $\pm$ 0.68\% & 39.30 $\pm$ 0.66\% &  53.44 $\pm$ 0.64\% & 31.59 $\pm$ 0.65\% & 41.95 $\pm$ 0.62\%& 28.46 $\pm$ 0.56\% & 39.88 $\pm$ 0.63\%\tabularnewline
		Proto Net~\cite{snell2017prototypical} & &&&&&&&\tabularnewline
		NCA~\cite{wu2018improving} & 60.08 $\pm$ 0.93\%  & 77.14 $\pm$ 0.72\% & 35.38 $\pm$ 0.70\% &  46.81 $\pm$ 0.72\%  & 44.07 $\pm$ 0.82\% &  58.28 $\pm$ 0.72\% & 25.04 $\pm$ 0.47\% & 31.24 $\pm$ 0.52\% \tabularnewline
	
		CosLoss & 57.22 $\pm$ 0.85\% & 75.34 $\pm$ 0.69\%& 39.60$\pm$ 0.70\%& 55.70$\pm$ 0.66\% &  43.16 $\pm$ 0.84\%& 59.32$\pm$ 0.77\%& 29.57 $\pm$ 0.70\% & 40.78 $\pm$ 0.68\%
		\tabularnewline
	
		Ours Eu & 48.26 $\pm$ 0.88\%    & 63.34 $\pm$ 0.74\%& \textbf{36.73 $\pm$ 0.69\%} & 50.96 $\pm$ 0.58\% & 37.48 $\pm$ 0.75\% & 47.35$\pm$ 0.70\% & 28.05 $\pm$ 0.53\% & 37.87  $\pm$ 0.62 \%  \tabularnewline
		
		Ours &  \textbf{61.0 $\pm$ 0.81\%}  &\textbf{75.60 $\pm$ 0.66\%}&  \textbf{40.16 $\pm$ 0.68\%} & \textbf{56.96 $\pm$ 0.65\%} & \textbf{37.33 $\pm$ 0.65\%} & \textbf{49.97 $\pm$ 0.66\%} & \textbf{31.20 $\pm$ 0.58\%} & \textbf{47.10 $\pm$ 0.62\%} \tabularnewline
		\bottomrule
	\end{tabular}}
	}
\end{table*}
\fi
		
%%%%%%%%% Generalizing

\subsection{Generalizing to Other Datasets}
\label{sec:generalization}
A new dataset may present data distribution shift, and the classification accuracy of widely used models drops significantly~\cite{recht2018cifar}.
In current setting of few-shot classification, most methods conduct training and testing phases within the same dataset, \ie, miniImageNet.
Although the training classes and testing classes do not share the same label space, they still comes from the same data distribution.
%, there are still lots of similarity between the training and testing label space.
%For example, different species of dogs may appear at
%\begin{figure}
%	\includegraphics[width=1.0\linewidth]{figs/discuss.pdf}
%	\caption{Blue box shows training example class. Yellow box shows testing example.}
%	\label{class diff}
%\end{figure}
%(\eg, \lqyu{A dog in training class and B dog in testing class}).
%
While, in the real world, the unknown novel classes may comes from an agnostic data distribution.
Therefore, to validate the generalization capability of our approach, we conduct the few-shot classification on novel classes from the following four datasets using the model trained on the miniImageNet training dataset.
\begin{itemize}
	\item
	\textbf{Caltech-101.}\
	The Caltech-101 dataset~\cite{fei2007learning,fei2006one} contains objects belonging to 101 categories. Each category contains about 40 to 800 images. Most categories have about 50 images.
	
	\item
	\textbf{Caltech-UCSD Birds-200-2011 (CUB-200).}\
	Caltech-UCSD Birds 200 (CUB-200)~\cite{WelinderEtal2010} contains photos of 200 bird species (mostly North American).
	In this fine-grained dataset, subtle differences between very similar classes can hardly be recognized even by humans.
	
	\item
	\textbf{Stanford Dogs.}\
	The Stanford Dogs dataset~\cite{KhoslaYaoJayadevaprakashFeiFei_FGVC2011} contains images of 120 breeds of dogs from around the world. This dataset has been built using images and annotation from ImageNet for the task of fine-grained image categorization.
	
	\item
	\textbf{Stanford Cars.}\
	The Stanford Cars~\cite{KrauseStarkDengFei-Fei_3DRR2013} contains 16,185 images of 196 classes of cars.
\end{itemize}

Following the same data selection principal as miniImageNet~\cite{vinyals2016matching}, we randomly select 20 classes in each dataset as the test dataset.
Note that the test datasets do not share the same label space with the training images.
Please see the section 1 in the supplementary files for detailed selected class in each dataset.
Without any fine-tuning, we directly use the model trained on the miniImageNet training dataset to perform few-shot classification on the new datasets.
Table~\ref{tab:caltech101} shows the classification performance of Relation Net, Proto Net and our method on the four datasets.
The results are achieved by the model with the same network backbone.
It is observed that our model performs consistently better than Relation Net and Proto Net on all four datasets.
To compare the results on the different datasets, the accuracy on Caltech-101 are much higher than the results of other three datasets, even than the miniImageNet testing dataset.
This is because the Caltech-101 contains a single object with pure background and it is much easier to be recognized, while the CUB-200, Stanford Dogs and Stanford Cars have relative complex background.
%
%We noticed that the results achieved on Stanford Cars are a bit lower than that on other datasets. This is because "Cars" is never been learned in the training dataset.\xm{Here may need rephase.}
%Even though, we still can achieve around 47.10\% results on 5-way 5-shot setting.
%
We visualize the results of 5-way 5-shot setting achieved by Relation Net and our model in Figure~\ref{fig:cub200}.
We can see that our method is very discriminative to similar objects.
These comparisons clearly demonstrate that our approach is able to learn more generalized transferable features for few-shot classification among different datasets.
Please see more visualized results in the section 3 in the supplementary files.

\subsection{Analysis of Our Method}
\label{sec:analysis}
To better understand our method, we conduct the following experiments on the miniImageNet dataset.

\subsubsection{Results with Different Network Backbones}
We compare the few-shot classification performance of our approach under different network backbones, \ie, AlexNet~\cite{krizhevsky2012imagenet}, VGG~\cite{Simonyan2015} and ResNet~\cite{he2016deep}.
We conduct experiments on the miniImageNet with the same experiment setting for network architectures.
Note that the performance is evaluated by one nearest neighborhood (1-NN) with Euclidean distance to better show the influence of different network backbones.
From the results in Table~\ref{tab:backone}, it is observed that the classification accuracy of AlexNet and VGG11 are similar, while the few-shot classification accuracy is largely improved (about 10\% improvement in both 5-way 1-shot and 5-way 5-shot settings) with a more deeper ResNet.
The reason may be that we can extract more representative features with the deeper ResNet and thus improve the accuracy on few-shot testing.
%%%
ResNet18 and ResNet34 achieve similar results on 5-way 5-shot evaluation, but ResNet34 achieves a bit higher performance on 5-way 1-shot setting.
However, the classification accuracy would be decreased as the model complexity continues to grow (\eg, from ResNet34 to ResNet50).
This finding indicates that too many parameters may lead to overfitting on the training tasks and thus decrease the classification results on novel categories.
Therefore, an effective network backbone can indeed contribute to the transferable feature extraction and improve the accuracy on few-shot classification.
Overall, in our experiment, we choose the ResNet34 as the network backbone.
%However, the performance would not improved as the model complexity grows.

\begin{table}[h]
	\centering
	\caption{Few-shot classification accuracy (\%) for 600 runs with 95\% confidence intervals with different network backbones on the miniImageNet testing data.}
\label{tab:backone}
{	
\resizebox{0.49\linewidth}{!}{\begin{tabular}{ccc}
		\hline
		\toprule
		\multirow{2}{*}{Backbone}& \multicolumn{2}{c}{ 5-way Acc.} \tabularnewline
		
		&  1-shot  & 5-shot  \tabularnewline
		\hline
		
		AlexNet & 44.78 $\pm$ 0.78 & 58.69 $\pm$ 0.71  \tabularnewline
		VGG 11 & 44.74 $\pm$ 0.81 & 58.55 $\pm$ 0.66 \tabularnewline
		
		ResNet18	&  53.62 $\pm$ 0.84 & \textbf{68.27 $\pm$ 0.67} \tabularnewline
		
		ResNet34	& \textbf{54.46 $\pm$ 0.89}  & 68.15 $\pm$ 0.65 \tabularnewline
		
		ResNet50	& 53.46 $\pm$ 0.88  &  65.32 $\pm$ 0.72 \tabularnewline
		
		\bottomrule
		\hline
	\end{tabular}}}
\end{table}

\begin{table}[!h]
	\centering
	\caption{\revise{Few-shot classification accuracy (\%) for 600 runs with 95\% confidence intervals on the miniImageNet testing data with different number of negatives in the tuplet loss .  }}
	\label{tab:negativeclasses} %
	{	
	\resizebox{0.5\linewidth}{!}{\begin{tabular}{ccc}
			\hline
			\toprule
			\multirow{2}{*}{Number of $K$}& \multicolumn{2}{c}{ 5-way Acc.} \tabularnewline
			
			&  1-shot  & 5-shot  \tabularnewline
			\hline
			$K$=1 (\emph{triplet loss})		&  40.15 $\pm$ 0.75  & 54.62 $\pm$ 0.68  \tabularnewline
			
			$K$=4 	&  51.22 $\pm$ 0.81 & 65.66 $\pm$ 0.68  \tabularnewline
			
			$K$=5 	&  \textbf{54.46 $\pm$ 0.89} & \textbf{68.15 $\pm$ 0.65}  \tabularnewline

			$K$=8		& 53.17 $\pm$ 0.81  & 66.77 $\pm$ 0.68 \tabularnewline
			
			$K$=16	& 46.03 $\pm$ 0.79  &  60.02 $\pm$ 0.67  \tabularnewline
			
			%	1-class 5-sample	&  46.38 $\pm$ 0.82\% & 57.24 $\pm$ 0.73\%  \tabularnewline
			
			%	4-class 5-sample	&  46.21 $\pm$ 0.83\%  & 61.91 $\pm$ 0.68\% \tabularnewline
			\bottomrule
			\hline
		\end{tabular}}}
\end{table}

\subsubsection{The Tuplet-loss with Different Negative Pairs}
We compare the performance of our method with different $K$ in the tuplet loss, where $K$ is the number of negative samples from different classes in each tuplet.
We also report the classification accuracy using the one nearest neighborhood (1-NN) classifier with Euclidean distance.
As shown in Table~\ref{tab:negativeclasses}, the accuracy is a little low if we set $K$ as 1 in the tuplet loss (equivalent to traditional triplet loss).
%
%This is conform with our argument that learning feature embedding with triplet loss is not very suitable for few-shot learning.
%
The classification accuracy is improved with a larger $K$, since the anchor sample interacts with more samples in one mini-batch and makes the gradient more stable.
In another aspect, the classification accuracy would be saturated with a bigger $K$, and we can achieve the best performance when setting $K$ to 5.

\revise{It is worth mentioning that when $K$ equal to 1, $K$-tuple loss is triplet loss. Compared with triplet loss, the classification accuracy is improved with $K$-tuplet loss, since the anchor sample interacts with more samples in one mini-batch and makes the gradient more stable. 
As shown in firs line in Table~\ref{tab:negativeclasses}, the performance of triplet loss ($K$ =1) on our few-shot learning task is 40.15\% and 54.62\% for 1-shot and 5-shot learning. While our $K$-tuplet loss can achieve 54.46\% and 68.15\% for 1-shot and 5-shot learning respectively, surpassing the triplet loss by around 14\% on both 1-shot and 5-shot setting. 
}
\subsubsection{Effects of Semi-hard Mining}
\revise{
Table~\ref{tab:semihard} shows the effects on our feature embedding when trained with and without semi-hard mining. We report the few-shot classification accuracy on the \emph{mini}ImageNet testing data with the two resulting learned feature embedding.
``w/o semi-hard'' denotes the model trained with equation~(\ref{eq1}) for 100 epochs while ``w semi-hard'' refers to model trained with equation~\ref{eq1} for 80 epochs and then utilize equation~(\ref{eq2}) for remaining 20 epochs. 
It is observed that with semi-hard mining, the few-shot classification accuracy on both 1-shot and 5-shot scenarios can be further improved by relative 1.6\% and 1.0\% respectively. This comparison demonstrates the effectiveness of ``semi-hard mining strategy'' to improve feature embedding learning.
}
\begin{table}[h]
	\centering
	\caption{\revise{The effects of semi-hard mining. The report results are averaged few-shot classification for 600 runs with 95\% confidence intervals (Unit: \%).}}
	\label{tab:semihard} %
	{		
	\resizebox{0.5\linewidth}{!}{\begin{tabular}{ccc}
			\hline
			\toprule
			\multirow{2}{*}{Setting}& \multicolumn{2}{c}{ 5-way acc} \tabularnewline
			&  1-shot  & 5-shot  \tabularnewline
			\hline
			w/o semi-hard	&  53.62 $\pm$ 0.84 & 67.48 $\pm$ 0.68 \tabularnewline
			w semi-hard	&  \textbf{54.46 $\pm$ 0.89} &  \textbf{68.15 $\pm$ 0.65}  \tabularnewline
			\bottomrule
			\hline
		\end{tabular}}
	}
\end{table}

\if 1
\subsubsection{Analysis on Feature Embedding}
We explore the importance of different embedding loss function in our few-shot learning framework.
We compare with the state-of-the-art feature embedding method, \ie, CosFace~\cite{wang2018cosface} on the face recognition.
The comparison conducted under the same network backbone, and employ the similarity module as well as the same training strategy.
%Figure~\ref{class diff} shows that our method performs much better than the CosFace feature embedding.

\begin{table}[!h]
	\centering
	\caption{Comparison with different embedding function for few-shot classification on the miniImageNet.}
	\label{tab:cosface}
	{\begin{tabular}{ccc}
	\toprule
	\multirow{2}{*}{setting}& \multicolumn{2}{c}{ 5-way acc} \tabularnewline
	
	&  1-shot  & 5-shot  \tabularnewline
	\hline
	CosFace	&    & 69.37 $\pm$ 0.68  \tabularnewline
	
	ours	&  \textbf{58.53 $\pm$ 0.82} & \textbf{72.57 $\pm$ 0.62}  \tabularnewline
	\bottomrule
\end{tabular}}
\end{table}
\fi
\subsubsection{The Analysis of Different Margin}
We also investigate the effect of different margin $\alpha$ in the tuplet loss and the results of our whole framework with different settings are shown in Table~\ref{tab:margin}.
The experimental results show that with margin 0.5, the feature embedding in this task is the best.
A smaller margin will decrease the performance due to the inter-class variation is not well-learned. And a larger margin may increase the difficulty in the training.

\begin{table}[t]
	\centering
	\caption{\revise{Few-shot classification accuracy (\%) for 600 runs with 95\% confidence intervals on the miniImageNet testing data with different margins.}}
	\label{tab:margin}
\resizebox{0.5\linewidth}{!}{	\begin{tabular}{ccc}
			\hline
			\toprule
			\multirow{2}{*}{Margin $\alpha$}& \multicolumn{2}{c}{ 5-way Acc.} \tabularnewline
			&  1-shot  & 5-shot  \tabularnewline
			\hline
			0.10	&  54.46 $\pm$ 0.89  & 68.15 $\pm$ 0.65 \tabularnewline
			
			0.30 & 57.10 $\pm$ 0.64   &   71.02 $\pm$ 0.58    \tabularnewline
			0.50	&  \textbf{58.30 $\pm$ 0.84} & \textbf{72.37 $\pm$ 0.63} \tabularnewline
			
			0.80    &        53.05 $\pm$ 0.41   & 62.02 $\pm$ 0.51
			\tabularnewline
			1.00	&  49.30 $\pm$ 0.07  & 59.62 $\pm$ 0.07
			\tabularnewline
			\bottomrule
			\hline
		\end{tabular}	}\end{table}

\begin{table}[t]
	\centering
	\caption{Averaged accuracy for 600 runs with 95\% confidence intervals on the miniImageNet testing data with additional training classes from ImageNet dataset. N is the the number of extra classes in the notation of ``64+N''.}
	\label{tab:additionaldataset}
	{	
		\resizebox{0.5\linewidth}{!}{\begin{tabular}{ccc}
				\hline
				\toprule
				\multirow{2}{*}{Training Class}& \multicolumn{2}{c}{ 5-way Acc.} \tabularnewline
				
				&  1-shot  & 5-shot  \tabularnewline
				\hline
				64 + 0	&  58.30 $\pm$ 0.84  & 72.37 $\pm$ 0.63 \tabularnewline
				
				64 + 64 &  61.12 $\pm$ 0.06 & 75.14 $\pm$ 0.67 \tabularnewline
				
				64 + 128  &  \textbf{65.60 $\pm$ 0.07}  & \textbf{77.74 $\pm$ 0.07} \tabularnewline
				
				\bottomrule
				\hline
			\end{tabular}}
		}	
	\end{table}

\subsubsection{The Results with More Training Classes}
We would like to explore whether the few-shot classification accuracy will increase if more training classes are available.
Thus, we conduct experiments with additional class images from the ImageNet dataset.
Note that the additional dataset does not share the same labels with the testing images.
Table~\ref{tab:additionaldataset} presents the accuracy on the miniImageNet testing dataset of our method trained with different number of training classes.
We can see that our method can be further improved with extra training classes data available.
This is conform with our expectation that we can learn more transferable generalized feature embedding from more training samples.
Based on the generalized feature, we can further improve the few-shot classification accuracy on the novel categories.

\begin{figure*}[h]
	\centering
	\includegraphics[width=1.0\linewidth]{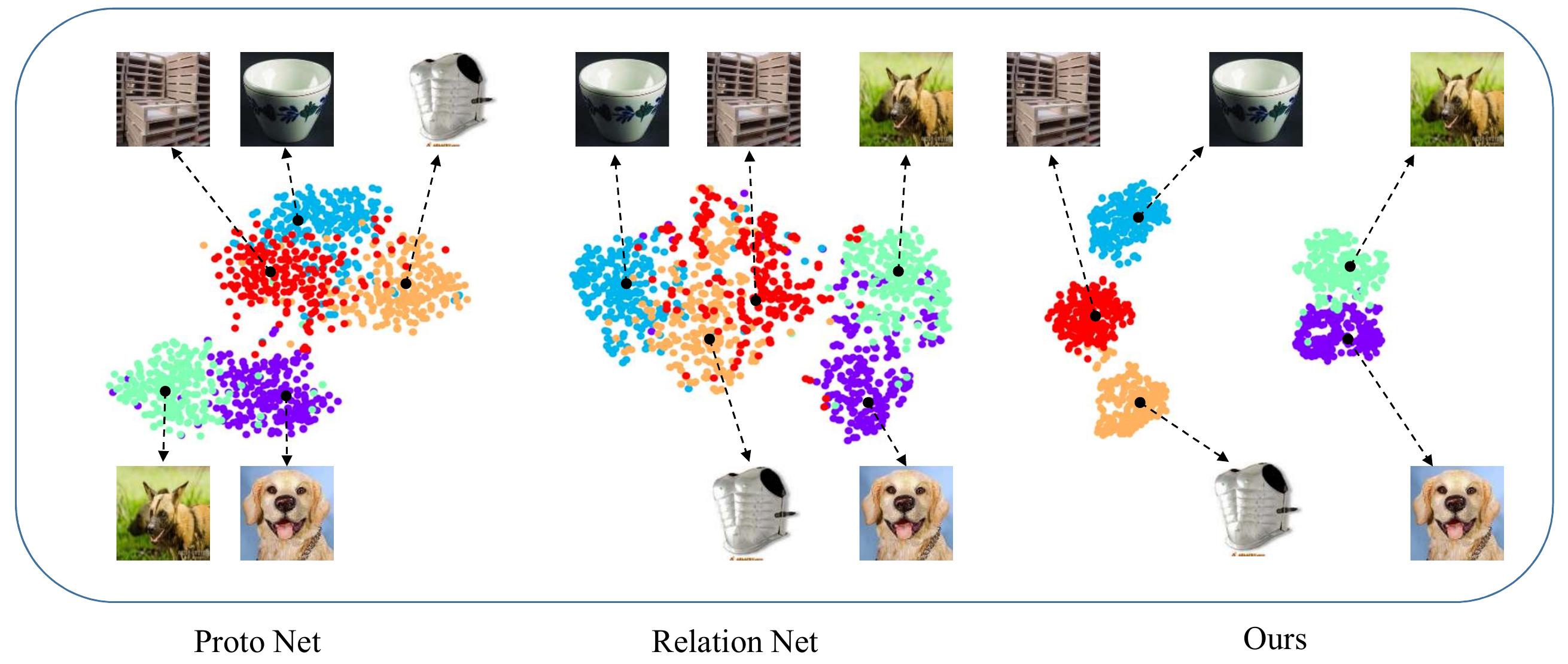}
	\caption{T-SNE visualization of features in Proto Net, Relation Net and our method on the same set of samples in the test dataset (example 1).}
	\label{fig:feature1}
\end{figure*}
\begin{figure*}[h]
	\centering
	\includegraphics[width=1.0\linewidth]{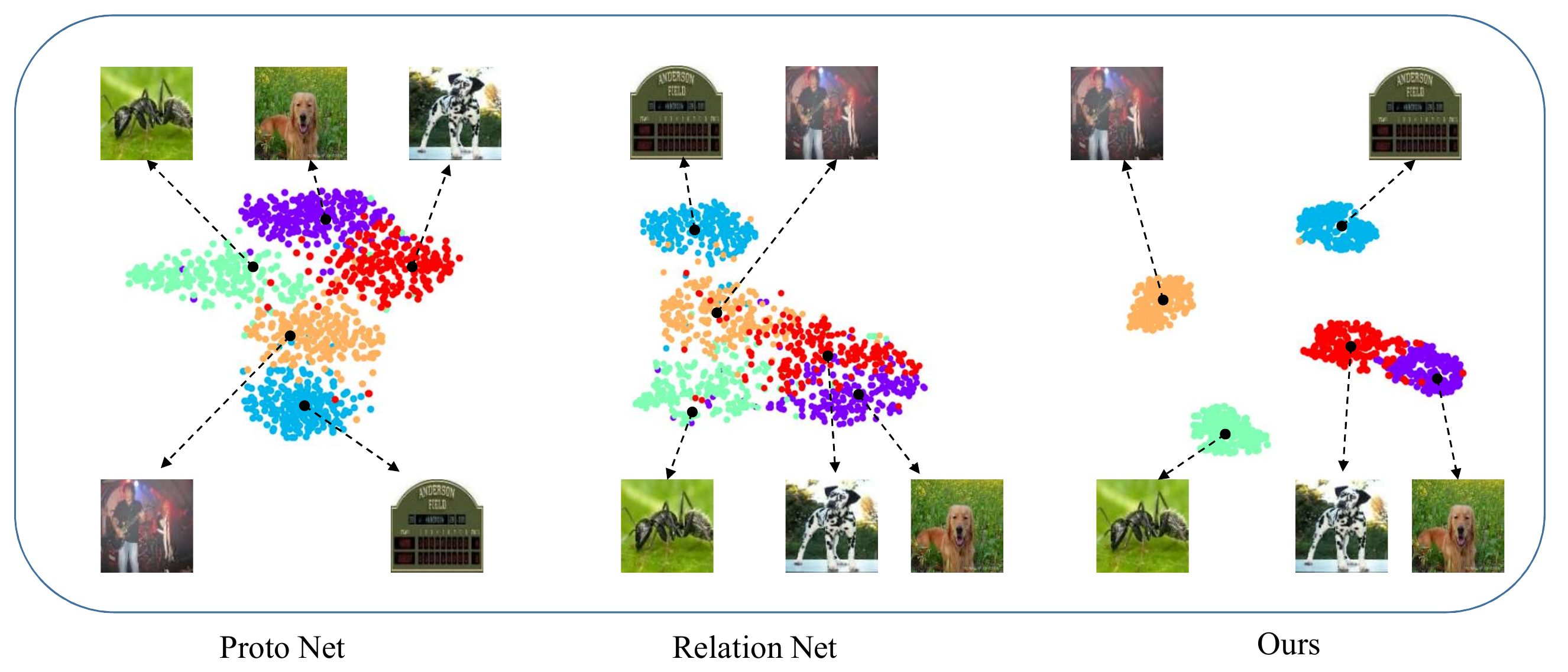}
	\caption{T-SNE visualization of features in Proto Net, Relation Net and our method on the same set of samples in the test dataset (example 2).}
	\label{fig:feature2}
\end{figure*}

\section{Comparison of Visualized Features}
The effectiveness of our method is mainly due to a well-learned feature embedding, which improves the few-shot classification performance on the \emph{novel classes}.
To show the generalization and discrimination of our learned feature embedding on novel class samples, we visualize the features in comparison with Proto Net~\cite{snell2017prototypical} and Relation Net~\cite{sung2018learning}; see Figures~\ref{fig:feature1} and~\ref{fig:feature2}.
The feature embedding is learned from the miniImageNet training dataset and tested on the miniImageNet test dataset.
For each figure, we mimic the test procedure by randomly selecting five classes from the test dataset. 
Then, we compute the features of 200 samples per class and create the visualizations of the features shown in each figure using t-SNE~\cite{maaten2008visualizing}, where we use the same 200 samples for different methods in each figure.

From Figure~\ref{fig:feature1}, we can see that our feature embedding can well separate the five classes, especially for cuirass, crate, and mixing bowl, as compared to Proto Net and Relation Net.
Although it is quite challenging to distinguish the two species of dogs shown in the unseen novel classes, our method can still better separate their features compared with Proto Net and Relation Net.
Since the feature embeddings are visualized on the novel classes, the results clearly demonstrate that our method produces better feature embeddings on the novel classes compared to the other two methods.
Therefore, our results make it easier for the subsequent K-NN to perform the classification, thus leading to more promising few-shot classification results.
Please see more visualized results in the section 2 in the supplementary file.

\section{Conclusion}
\revise{
In this work, we revisit the metric learning and propose a simple and effective $K$-tuplet network for few-shot
learning. 
We present an efficient $K$-tuplet network to utilize the relationship of training samples to learn the transferable feature embedding that performs well not only on the training samples but also on the novel class samples. }
Built on top of this generalized feature embedding, we can largely improve the few-shot classification accuracy.
%We also verify that the image-to-class measure is superior to the image-to-image measure, owing to the exchangeability of visual patterns within a class
% we explore the importance of feature embedding in the few-shot learning, an important topic but neglect in the current work. 
Our method is simple yet effective, and outperforms other metric-based few-shot classification algorithms on the public benchmark dataset.  
More importantly, our method can generalize very well to the novel categories even on other four datasets.

\if 1 
\section{Acknowledgments}
This work is supported by a grant from Key-Area Research and Development Program of Guangdong Province, China (2020B010165004)
and a grant from National Natural Science Foundation of China with Project No. U1813204.
\fi 
%\IEEEpeerreviewmaketitle

%\bibliographystyle{IEEEtran}
%\bibliography{IEEEabrv,bibliography}
%\bibliographystyle{IEEEtran}
%\bibliography{refs}

\bibliographystyle{elsarticle-num}

\bibliography{refs}

\begin{thebibliography}{10}
\expandafter\ifx\csname url\endcsname\relax
  \def\url#1{\texttt{#1}}\fi
\expandafter\ifx\csname urlprefix\endcsname\relax\def\urlprefix{URL }\fi
\expandafter\ifx\csname href\endcsname\relax
  \def\href#1#2{#2} \def\path#1{#1}\fi

\bibitem{Czhang2019}
C.~{Zhang}, C.~{Li}, J.~{Cheng}, Few-shot visual classification using image
  pairs with binary transformation, IEEE Transactions on Circuits and Systems
  for Video Technology\href {http://dx.doi.org/10.1109/TCSVT.2019.2920783}
  {\path{doi:10.1109/TCSVT.2019.2920783}}.

\bibitem{snell2017prototypical}
J.~Snell, K.~Swersky, R.~Zemel, Prototypical networks for few-shot learning,
  in: NIPS, 2017, pp. 4077--4087.

\bibitem{finn2017model}
C.~Finn, P.~Abbeel, S.~Levine, Model-agnostic meta-learning for fast adaptation
  of deep networks, in: ICML, 2017, pp. 1126--1135.

\bibitem{Sachin2017}
S.~Ravi, H.~Larochelle, Optimization as a model for few-shot learning, in:
  ICLR, 2017.

\bibitem{wertheimer2019few}
D.~Wertheimer, B.~Hariharan, Few-shot learning with localization in realistic
  settings, in: CVPR, 2019, pp. 6558--6567.

\bibitem{li2019revisiting}
W.~Li, L.~Wang, J.~Xu, J.~Huo, Y.~Gao, J.~Luo, Revisiting local descriptor
  based image-to-class measure for few-shot learning, in: CVPR, 2019.

\bibitem{chen2019closer}
W.-Y. Chen, Y.-C. Liu, Z.~Kira, Y.-C.~F. Wang, J.-B. Huang, A closer look at
  few-shot classification, in: ICLR, 2019.

\bibitem{bertinetto2018metalearning}
L.~Bertinetto, J.~F. Henriques, P.~Torr, A.~Vedaldi, Meta-learning with
  differentiable closed-form solvers, in: ICLR, 2019.

\bibitem{jamal2018task}
M.~A. Jamal, G.-J. Qi, M.~Shah, Task-agnostic meta-learning for few-shot
  learning, in: CVPR, 2019.

\bibitem{li2019ctm}
H.~Li, D.~Eigen, S.~Dodge, M.~Zeiler, X.~Wang, Finding task-relevant features
  for few-shot learning by category traversal, in: CVPR, 2019.

\bibitem{lee2019meta}
K.~Lee, S.~Maji, A.~Ravichandran, S.~Soatto, Meta-learning with differentiable
  convex optimization, in: Proceedings of the IEEE Conference on Computer
  Vision and Pattern Recognition, 2019, pp. 10657--10665.

\bibitem{kim2019edge}
J.~Kim, T.~Kim, S.~Kim, C.~D. Yoo, Edge-labeling graph neural network for
  few-shot learning, in: CVPR, 2019, pp. 11--20.

\bibitem{liu2019large}
Z.~Liu, Z.~Miao, X.~Zhan, J.~Wang, B.~Gong, S.~X. Yu, Large-scale long-tailed
  recognition in an open world, in: CVPR, 2019, pp. 2537--2546.

\bibitem{lifchitz2019dense}
Y.~Lifchitz, Y.~Avrithis, S.~Picard, A.~Bursuc, Dense classification and
  implanting for few-shot learning, in: CVPR, 2019.

\bibitem{liu2018learning}
Y.~Liu, J.~Lee, M.~Park, S.~Kim, E.~Yang, S.~J. Hwang, Y.~Yang, Learning to
  propagate labels: Transductive propagation network for few-shot learning, in:
  ICLR, 2018.

\bibitem{sun2019meta}
Q.~Sun, Y.~Liu, T.-S. Chua, B.~Schiele, Meta-transfer learning for few-shot
  learning, in: CVPR, 2019, pp. 403--412.

\bibitem{cheny2019multi}
Z.~Chen, Y.~Fuy, Y.~Zhang, Y.-G. Jiang, X.~Xue, L.~Sigal, Multi-level semantic
  feature augmentation for one-shot learning, IEEE Transactions on Image
  Processing.

\bibitem{chen2019imageblock}
Z.~Chen, Y.~Fu23, K.~Chen, Y.-G. Jiang123, Image block augmentation for
  one-shot learning, Vol.~6, 2019.

\bibitem{chen2019imagedeform}
Z.~Chen, Y.~Fu, Y.-X. Wang, L.~Ma, W.~Liu, M.~Hebert, Image deformation
  meta-networks for one-shot learning, in: CVPR, 2019, pp. 8680--8689.

\bibitem{long2018object}
L.~Long, W.~Wang, J.~Wen, M.~Zhang, Q.~Lin, B.~C. Ooi, Object-level
  representation learning for few-shot image classification, arXiv:1805.10777.

\bibitem{schwartz2018delta}
E.~Schwartz, L.~Karlinsky, J.~Shtok, S.~Harary, M.~Marder, R.~Feris, A.~Kumar,
  R.~Giryes, A.~M. Bronstein, Delta-encoder: an effective sample synthesis
  method for few-shot object recognition, arXiv:1806.04734.

\bibitem{oreshkin2018tadam}
B.~N. Oreshkin, A.~Lacoste, P.~Rodriguez, Tadam: Task dependent adaptive metric
  for improved few-shot learning, arXiv:1805.10123.

\bibitem{li2019large}
A.~Li, T.~Luo, Z.~Lu, T.~Xiang, L.~Wang, Large-scale few-shot learning:
  Knowledge transfer with class hierarchy, in: CVPR, 2019, pp. 7212--7220.

\bibitem{liu2019lcc}
Y.~Liu, Q.~Sun, A.-A. Liu, Y.~Su, B.~Schiele, T.-S. Chua, Lcc: Learning to
  customize and combine neural networks for few-shot learning, in: CVPR, 2019.

\bibitem{zhang2016contextual}
C.~Zhang, Q.~Huang, Q.~Tian, Contextual exemplar classifier-based image
  representation for classification, IEEE Transactions on Circuits and Systems
  for Video Technology 27~(8) (2016) 1691--1699.

\bibitem{jamal2019task}
M.~A. Jamal, G.-J. Qi, Task agnostic meta-learning for few-shot learning, in:
  Proceedings of the IEEE Conference on Computer Vision and Pattern
  Recognition, 2019, pp. 11719--11727.

\bibitem{srivastava2014dropout}
N.~Srivastava, G.~Hinton, A.~Krizhevsky, I.~Sutskever, R.~Salakhutdinov,
  Dropout: a simple way to prevent neural networks from overfitting, The
  Journal of Machine Learning Research 15~(1) (2014) 1929--1958.

\bibitem{lee2015deeply}
C.-Y. Lee, S.~Xie, P.~Gallagher, Z.~Zhang, Z.~Tu, Deeply-supervised nets, in:
  Artificial Intelligence and Statistics, 2015, pp. 562--570.

\bibitem{santoro2016meta}
A.~Santoro, S.~Bartunov, M.~Botvinick, D.~Wierstra, T.~Lillicrap, Meta-learning
  with memory-augmented neural networks, in: ICML, 2016, pp. 1842--1850.

\bibitem{munkhdalai2017meta}
T.~Munkhdalai, H.~Yu, Meta networks, in: ICML, 2017, pp. 2554--2563.

\bibitem{koch2015siamese}
G.~Koch, R.~Zemel, R.~Salakhutdinov, Siamese neural networks for one-shot image
  recognition, in: ICML Deep Learning Workshop, Vol.~2, 2015.

\bibitem{vinyals2016matching}
O.~Vinyals, C.~Blundell, T.~Lillicrap, D.~Wierstra, et~al., Matching networks
  for one shot learning, in: NIPS, 2016, pp. 3630--3638.

\bibitem{sung2018learning}
F.~Sung, Y.~Yang, L.~Zhang, T.~Xiang, P.~H. Torr, T.~M. Hospedales, Learning to
  compare: Relation network for few-shot learning, in: CVPR, 2018.

\bibitem{fei2006one}
L.~Fei-Fei, R.~Fergus, P.~Perona, One-shot learning of object categories, IEEE
  Transactions on Pattern Analysis and Machine Intelligence 28~(4) (2006)
  594--611.

\bibitem{lake2011one}
B.~Lake, R.~Salakhutdinov, J.~Gross, J.~Tenenbaum, One shot learning of simple
  visual concepts, in: Proceedings of the Annual Meeting of the Cognitive
  Science Society, Vol.~33, 2011.

\bibitem{rusu2018meta}
A.~A. Rusu, D.~Rao, J.~Sygnowski, O.~Vinyals, R.~Pascanu, S.~Osindero,
  R.~Hadsell, Meta-learning with latent embedding optimization, ICLR.

\bibitem{mishra2018simple}
N.~Mishra, M.~Rohaninejad, X.~Chen, P.~Abbeel, A simple neural attentive
  meta-learner, in: ICLR, 2018.

\bibitem{gidaris2018dynamic}
S.~Gidaris, N.~Komodakis, Dynamic few-shot visual learning without forgetting,
  in: Proceedings of the IEEE Conference on Computer Vision and Pattern
  Recognition, 2018, pp. 4367--4375.

\bibitem{Act2Param}
S.~Qiao, C.~Liu, W.~Shen, A.~L. Yuille, Few-shot image recognition by
  predicting parameters from activations, in: CVPR, 2018.

\bibitem{bellet2013survey}
A.~Bellet, A.~Habrard, M.~Sebban, A survey on metric learning for feature
  vectors and structured data, arXiv preprint arXiv:1306.6709.

\bibitem{hadsell2006dimensionality}
R.~Hadsell, S.~Chopra, Y.~LeCun, Dimensionality reduction by learning an
  invariant mapping, in: CVPR, 2006.

\bibitem{hoffer2015deep}
E.~Hoffer, N.~Ailon, Deep metric learning using triplet network, in:
  International Workshop on Similarity-Based Pattern Recognition, Springer,
  2015, pp. 84--92.

\bibitem{sohn2016improved}
K.~Sohn, Improved deep metric learning with multi-class n-pair loss objective,
  in: NIPS, 2016, pp. 1857--1865.

\bibitem{liu2018transductive}
Y.~Liu, J.~Lee, M.~Park, S.~Kim, Y.~Yang, Transductive propagation network for
  few-shot learning, in: ICLR, 2019.

\bibitem{chopra2005learning}
S.~Chopra, R.~Hadsell, Y.~LeCun, Learning a similarity metric discriminatively,
  with application to face verification, in: CVPR, Vol.~1, 2005.

\bibitem{wang2014learning}
J.~Wang, Y.~Song, T.~Leung, C.~Rosenberg, J.~Wang, J.~Philbin, B.~Chen, Y.~Wu,
  Learning fine-grained image similarity with deep ranking, in: CVPR, 2014.

\bibitem{schroff2015facenet}
F.~Schroff, D.~Kalenichenko, J.~Philbin, Facenet: A unified embedding for face
  recognition and clustering, in: CVPR, 2015.

\bibitem{taigman2014deepface}
Y.~Taigman, M.~Yang, M.~Ranzato, L.~Wolf, Deepface: Closing the gap to
  human-level performance in face verification, in: CVPR, 2014.

\bibitem{lu2017discriminative}
J.~Lu, J.~Hu, Y.-P. Tan, Discriminative deep metric learning for face and
  kinship verification, IEEE Transactions on Image Processing 26~(9) (2017)
  4269--4282.

\bibitem{hu2017sharable}
J.~Hu, J.~Lu, Y.-P. Tan, Sharable and individual multi-view metric learning,
  IEEE transactions on pattern analysis and machine intelligence 40~(9) (2017)
  2281--2288.

\bibitem{xiao2017joint}
T.~Xiao, S.~Li, B.~Wang, L.~Lin, X.~Wang, Joint detection and identification
  feature learning for person search, in: CVPR, 2017.

\bibitem{Duan2020}
Y.~{Duan}, J.~{Lu}, W.~{Zheng}, J.~{Zhou}, Deep adversarial metric learning,
  IEEE Transactions on Image Processing 29 (2020) 2037--2051.

\bibitem{wu2018improving}
Z.~Wu, A.~A. Efros, X.~Y. Stella, Improving generalization via scalable
  neighborhood component analysis, in: ECCV, Springer, 2018, pp. 712--728.

\bibitem{ren2018meta}
M.~Ren, E.~Triantafillou, S.~Ravi, J.~Snell, K.~Swersky, J.~B. Tenenbaum,
  H.~Larochelle, R.~S. Zemel, Meta-learning for semi-supervised few-shot
  classification, in: ICLR, 2018.

\bibitem{garcia2017few}
V.~Garcia, J.~Bruna, Few-shot learning with graph neural networks, in: ICLR,
  2018.

\bibitem{li2019lgm}
H.~Li, W.~Dong, X.~Mei, C.~Ma, F.~Huang, B.-G. Hu, Lgm-net: Learning to
  generate matching networks for few-shot learning, in: ICML, 2019.

\bibitem{mehrotra2017generative}
A.~Mehrotra, A.~Dukkipati, Generative adversarial residual pairwise networks
  for one shot learning, arXiv:1703.08033.

\bibitem{ren18fewshotssl}
M.~Ren, E.~Triantafillou, S.~Ravi, J.~Snell, K.~Swersky, J.~B. Tenenbaum,
  H.~Larochelle, R.~S. Zemel, Meta-learning for semi-supervised few-shot
  classification, in: ICLR, 2018.

\bibitem{li2019finding}
H.~Li, D.~Eigen, S.~Dodge, M.~Zeiler, X.~Wang, Finding task-relevant features
  for few-shot learning by category traversal, in: Proceedings of the IEEE
  Conference on Computer Vision and Pattern Recognition, 2019, pp. 1--10.

\bibitem{he2016deep}
K.~He, X.~Zhang, S.~Ren, J.~Sun, Deep residual learning for image recognition,
  in: CVPR, 2016.

\bibitem{kingma2014adam}
D.~P. Kingma, J.~Ba, Adam: A method for stochastic optimization, in: ICLR,
  2015.

\bibitem{russakovsky2015imagenet}
O.~Russakovsky, J.~Deng, H.~Su, J.~Krause, S.~Satheesh, S.~Ma, Z.~Huang,
  A.~Karpathy, A.~Khosla, M.~Bernstein, et~al., Imagenet large scale visual
  recognition challenge, International Journal of Computer Vision 115~(3)
  (2015) 211--252.

\bibitem{munkhdalai2017rapid}
T.~Munkhdalai, X.~Yuan, S.~Mehri, A.~Trischler, Rapid adaptation with
  conditionally shifted neurons, in: ICML, 2018.

\bibitem{ye2018deep}
M.~Ye, Y.~Guo, Deep triplet ranking networks for one-shot recognition,
  arXiv:1804.07275.

\bibitem{wang2018large}
Y.~Wang, X.-M. Wu, Q.~Li, J.~Gu, W.~Xiang, L.~Zhang, V.~O. Li, Large margin
  few-shot learning, arXiv:1807.02872.

\bibitem{wang2018cosface}
H.~Wang, Y.~Wang, Z.~Zhou, X.~Ji, D.~Gong, J.~Zhou, Z.~Li, W.~Liu, Cosface:
  Large margin cosine loss for deep face recognition, in: CVPR, 2018, pp.
  5265--5274.

\bibitem{recht2018cifar}
B.~Recht, R.~Roelofs, L.~Schmidt, V.~Shankar, Do cifar-10 classifiers
  generalize to cifar-10?, arXiv preprint arXiv:1806.00451.

\bibitem{fei2007learning}
L.~Fei-Fei, R.~Fergus, P.~Perona, Learning generative visual models from few
  training examples: An incremental bayesian approach tested on 101 object
  categories, Computer vision and Image understanding 106~(1) (2007) 59--70.

\bibitem{WelinderEtal2010}
P.~Welinder, S.~Branson, T.~Mita, C.~Wah, F.~Schroff, S.~Belongie, P.~Perona,
  {Caltech-UCSD Birds 200}, Tech. Rep. CNS-TR-2010-001, California Institute of
  Technology (2010).

\bibitem{KhoslaYaoJayadevaprakashFeiFei_FGVC2011}
A.~Khosla, N.~Jayadevaprakash, B.~Yao, L.~Fei-Fei, Novel dataset for
  fine-grained image categorization, in: First Workshop on Fine-Grained Visual
  Categorization, CVPR, Colorado Springs, CO, 2011.

\bibitem{KrauseStarkDengFei-Fei_3DRR2013}
J.~Krause, M.~Stark, J.~Deng, L.~Fei-Fei, 3d object representations for
  fine-grained categorization, in: IEEE Workshop on 3D Representation and
  Recognition (3dRR-13), Sydney, Australia, 2013.

\bibitem{krizhevsky2012imagenet}
A.~Krizhevsky, I.~Sutskever, G.~E. Hinton, Imagenet classification with deep
  convolutional neural networks, in: NIPS, 2012, pp. 1097--1105.

\bibitem{Simonyan2015}
K.~Simonyan, A.~Zisserman, Very deep convolutional networks for large-scale
  image recognition, in: ICLR, 2015.

\bibitem{maaten2008visualizing}
L.~v.~d. Maaten, G.~Hinton, Visualizing data using t-sne, Journal of machine
  learning research 9~(Nov) (2008) 2579--2605.

\end{thebibliography}

\if 1 
\section*{Author Biographies}
\parpic{\includegraphics[width=1in,height=1.5in,clip,keepaspectratio]{photo/meng3.jpg}}

\noindent \textbf{Xiaomeng Li} is a postdoctoral research fellow in the Department of Radiation Oncology at Stanford University. Before that, she obtained Ph.D. degree in the Department of Computer Science and Engineering, The Chinese University of Hong Kong (CUHK) in July 2019. She received her B. Eng degree from Xidian University.
Her research interests include computer vision, deep learning and various applications in the medical image domain.

\parpic{\includegraphics[width=1in,height=1.5in,clip,keepaspectratio]{photo/quan.jpg}}
\noindent \textbf{Lequan Yu} received the B. Eng degree from Department of Computer Science and Technology in Zhejiang University in 2015.
He obtained Ph.D degree in the Department of Computer Science and Engineering, The Chinese University of Hong Kong (CUHK) in July 2019 .
He is currently a postdoctoral research fellow in the Department of Radiation Oncology at Stanford University.
His research lies at the intersection of medical image analysis and artificial intelligence.

\parpic{\includegraphics[width=1in,height=1.5in,clip,keepaspectratio]{photo/philip.png}}
\noindent \textbf{Chi-Wing Fu} is currently an associate professor in the Chinese University of Hong Kong.  He served as the co-chair of SIGGRAPH ASIA 2016's Technical Brief and Poster program, associate editor of Computer Graphics Forum, and panel member in SIGGRAPH 2019 Doctoral Consortium, as well as program committee members in various research conferences, including SIGGRAPH Asia Technical Brief, SIGGRAPH Asia Emerging tech., IEEE visualization, CVPR, IEEE VR, VRST, Pacific Graphics, GMP, etc.  His recent research interests include computation fabrication, 3D computer vision, user interaction, and data visualization.

\parpic{\includegraphics[width=1in,height=1.3in,clip,keepaspectratio]{photo/fang.pdf}}
\noindent \textbf{Meng Fang} is a senior researcher at Tencent AI Lab.  His research focuses on Machine Learning and Natural Language Processing. Before he joined in Tencent AI Lab, he was a research fellow at the School of Computing and Information Systems, University of Melbourne. He obtained his Ph.D. from the University of Technology, Sydney, Australia.
Previously, he also worked as an intern at the Commonwealth Scientific and Industrial Research Organisation (CSIRO) and Microsoft Research Asia (MSRA).

\if 1 
\parpic{\includegraphics[width=1in,height=1.5in,clip,keepaspectratio]{photo/lei.jpg}}
\noindent \textbf{Lei Xing} is currently the Jacob Haimson Professor of Medical Physics and Director of Medical Physics Division of Radiation Oncology Department at Stanford University.  Dr. Xing obtained his PhD in Physics from the Johns Hopkins University in 1992. He has been a member of the Radiation Oncology faculty at Stanford since 1997. His research has been focused on inverse treatment planning, artificial intelligence (AI) in medicine, tomographic image reconstruction, CT, optical and PET imaging instrumentations, image guided interventions, nanomedicine, and applications of molecular imaging in radiation oncology. 
\fi 

\parpic{\includegraphics[width=1in,height=1.5in,clip,keepaspectratio]{photo/heng.png}}
\noindent \textbf{Pheng-Ann Heng} received his B.Sc. (Computer Science) from the National University of Singapore in 1985. 
He received his M.Sc. (Computer Science), M. Art (Applied Math) and Ph.D. (Computer Science) all from the Indiana University in 1987, 1988, 1992 respectively.
He is a professor at the Department of Computer Science and Engineering at The Chinese University of Hong Kong. He has served as the Department Chairman from 2014 to 2017 and as the Head of Graduate Division from 2005 to 2008 and then again from 2011 to 2016.
%He has served as the Director of Virtual Reality, Visualization and Imaging Research Center at CUHK since 1999. He has served as the Director of Center for Human-Computer Interaction at Shenzhen Institutes of Advanced Technology, Chinese Academy of Sciences since 2006. He has been appointed by China Ministry of Education as a Cheung Kong Scholar Chair Professor in 2007. 
His research interests include AI and VR for medical applications, surgical simulation, visualization, graphics and human-computer interaction.
\fi


%\begin{thebibliography}{00}


%\bibitem{}

%\end{thebibliography}
\end{document}